\documentclass[11pt]{article}

\usepackage[preprint]{acl}

\usepackage{times}
\usepackage{latexsym}

\usepackage{amsmath, amsthm, amssymb}
\usepackage[linesnumbered,ruled]{algorithm2e}
\usepackage{enumitem}
\usepackage{multirow, multicol}
\usepackage{booktabs}
\usepackage{makecell}
\usepackage{colortbl}
\usepackage{pifont}

\newtheorem{theorem}{Theorem}

\newtheorem{remark}{Remark}

\usepackage[T1]{fontenc}
\usepackage[utf8]{inputenc}

\usepackage{microtype}

\usepackage{inconsolata}

\usepackage{graphicx}

\title{\textsc{XMark}: Reliable Multi-Bit Watermarking for LLM-Generated Texts}

\author{
 \textbf{Jiahao Xu\textsuperscript{1,2}}
 \thanks{This work was done during Jiahao's internship at the Oak~Ridge~National~Laboratory.}
 \quad 
 \textbf{Rui Hu\textsuperscript{1}} \quad 
 \textbf{Olivera Kotevska\textsuperscript{2}} \quad 
 \textbf{Zikai Zhang\textsuperscript{1}}
\\
\\
 \textsuperscript{1}University of Nevada, Reno \\ 
 \textsuperscript{2}Oak Ridge National Laboratory
\\
 \texttt{\{jiahaox, ruihu, zikaiz\}@unr.edu} \quad \texttt{kotevskao@ornl.gov}
}

\begin{document}
\maketitle
\begin{abstract}
Multi-bit watermarking has emerged as a promising solution for embedding imperceptible binary messages into Large Language Model (LLM)-generated text, enabling reliable attribution and tracing of malicious usage of LLMs. Despite recent progress, existing methods still face key limitations: some become computationally infeasible for large messages, while others suffer from a poor trade-off between text quality and decoding accuracy. Moreover, the decoding accuracy of existing methods drops significantly when the number of tokens in the generated text is limited, a condition that frequently arises in practical usage. 
To address these challenges, we propose \textsc{XMark}, a novel method for encoding and decoding binary messages in LLM-generated texts. The unique design of \textsc{XMark}'s encoder produces a less distorted logit distribution for watermarked token generation, preserving text quality, and also enables its tailored decoder to reliably recover the encoded message with limited tokens. Extensive experiments across diverse downstream tasks show that \textsc{XMark} significantly improves decoding accuracy while preserving the quality of watermarked text, outperforming prior methods. The code is at \url{https://github.com/JiiahaoXU/XMark}.
\end{abstract}

\section{Introduction}
The rapid advancement and widespread adoption of Large Language Models (LLMs), both closed-source (e.g., ChatGPT~\cite{schulman2022chatgpt}) and open-source (e.g., LLaMA~\cite{llama2}), have endowed them with remarkable abilities to generate high-quality text. As a result, they have become integral to many text generation applications, such as question answering~\cite{perkins2023academic}. However, these powerful generative capabilities also raise significant security and ethical concerns, as malicious users can exploit LLMs to generate harmful content such as fake news, phishing emails, and fraudulent reviews~\cite{zhang2025character}. 

Recently, researchers have proposed watermarking methods that embed identifiable signals into text, enabling post hoc detection of AI-generated content, i.e., zero-bit watermarking~\cite{liu2024survey, wu2025survey, pan2024markllm}. However, zero-bit watermarking is insufficient to prevent misuse of AI. To address this limitation, multi-bit watermarking methods have been actively studied, which embed and extract binary messages within text. Such messages can convey richer information, like user IDs, timestamps, and other metadata~\cite{stealthink}.

Multi-bit watermarking methods can be broadly categorized into two types: (1) \textit{Distortion-free methods}, where the watermarked text follows the same logit distribution as the unwatermarked text~\cite{boroujeny2024multi, hu2024unbiased, sta, christ2024undetectable, kuditipudi2024robust, dathathri2024scalable, wu2024a}. (2) \textit{Logit-perturbation methods} which instead encode messages by perturbing the logits of selected tokens. Compared with distortion-free methods, they can embed richer watermark information and produce watermarked text that is more robust to text editing~\cite{stealthink}. These methods generally follow a common paradigm: at each token generation step of LLM, once the model logits are obtained, a hash seed, computed from a hash key and previously-generated token(s), is used to permute the model’s vocabulary. From this permuted vocabulary, a subset of tokens (the green list) is selected, and their logits are boosted to increase their sampling probability. The model then samples the next token from the modified logits. 
% a hash function takes as input a hash key and the previously generated token(s) and outputs a hash seed. This seed is then used to derive a permutation over the model’s vocabulary. Based on a method-specific strategy, a green list is constructed to determine which tokens should be perturbed, thereby increasing their sampling probability. 
During decoding, green-list tokens appear more frequently in the watermarked text, thereby providing the watermark signal needed to recover the encoded message~\cite{kgw, mpac, rsbh, depthw, cycleshift, ctwl, majormark}.

Early methods such as CycleShift~\cite{cycleshift}, CTWL~\cite{ctwl}, and DepthW~\cite{depthw} take the message to be encoded as input to the hash function, yielding a distinct hash seed for vocabulary permutation. However, their decoding process requires brute-force enumeration over all possible message candidates to identify the encoded one, which becomes computationally infeasible for longer messages. To address this, MPAC~\cite{mpac} introduces a block-wise method that divides the message into multiple blocks and encodes/decodes one block per token. Nonetheless, its encoder suffers from degraded text quality because it heavily constrains the size of the green list, causing noticeable distortion in token sampling probabilities. A recent method, StealthInk~\cite{stealthink}, improves text quality by directly boosting the sampling probabilities of tokens with larger logits while eliminating the chance of sampling tokens with smaller logits, thereby better preserving quality. However, this comes at the cost of weakening the watermark signal and decreasing decoding accuracy. Importantly, we observe that all existing methods rely on the availability of a sufficient number of tokens in the suspect text for reliable decoding. In practice, however, the length of the suspect text may be limited, which can cause a significant drop in decoding accuracy.

In this work, we propose \textsc{XMark}, which leverages green lists a\textit{cross} (\underline{\textsc{X}}) multiple vocabulary permutations to improve the text quality and decoding accuracy of multi-bit water\underline{\textsc{Mark}}ing. 
% that preserves text quality and achieves high decoding accuracy under limited-token scenarios.
%
% In this work, we propose \textsc{XMark}, a novel multi-bit watermarking method based on our insight of \textit{\underline{\textbf{cross (X)}}-permutation asymmetry} of each token during water\underline{\textbf{\textit{mark}}}ing. 
Specifically, \textsc{XMark} follows the block-wise encoding and decoding scheme as in~\cite{mpac}. 
% It is built on a new watermarking paradigm \textsc{LoSo} to construct a large green list during encoding. 
The core innovation of \textsc{XMark}'s encoding process is its use of $k$ distinct hash keys to generate $k$ permutations of the vocabulary. Each permutation is partitioned into $2^d$ shards, where $d$ is the length of a message block. For each permutation, a green list is formed by unioning all shards except the one indexed by the decimal value of the message block to be encoded, a paradigm we refer to as Leave-one-Shard-out (\textsc{LoSo}). An \textit{evergreen} list is constructed by intersecting all $k$ green lists, and a positive bias is added to its tokens’ logits to boost their probabilities to be sampled. This method ensures the size of the evergreen list is proportional to $(1-2^{-d})^k$ of the vocabulary, thereby largely preserving the quality of watermarked texts.

A novel decoder is designed with a \textit{constrained token–shard mapping matrix} (cTMM) to enhance decoding accuracy, particularly when only a limited number of generated tokens are available. 
For each token, the decoder reconstructs the same $k$ permutations and their corresponding shard partitions, incrementing by at most one the count of the shard that the token belongs to in the cTMM. 
Through this process, each token contributes to updating the cTMM up to $k$ times, providing a more accurate estimation of the mapping between a token and its originating shard and amplifying the distinction between the boosted shards (from which tokens are more likely drawn) and the unboosted shard (which encodes the message).  
% Tokens in the evergreen list, being positively perturbed during generation, are more likely to be sampled and consistently contribute to perturbed shards. 
% 
This design effectively enhances token–shard mapping construction and hence improves decoding reliability under limited-token conditions.
Extensive evaluations demonstrate that \textsc{XMark} consistently achieves higher decoding accuracy while maintaining text quality comparable to existing approaches, especially under limited-token settings.

\section{
% Preliminaries and 
Preliminaries and Background}

\subsection{Problem Formulation}

% \textbf{Problem Formulation.}
We consider the widely studied multi-bit watermarking setting~\cite{mpac, rsbh, majormark}, where a cloud-hosted LLM-as-a-Service provider stamps each model-generated response with a watermark. 
Given an LLM $f$, a user prompt $\mathbf{x}_p$, and a $b$\footnote{Throughout, we assume $b>0$ and is an even number.}-bit binary message $\mathbf{m}\in\{0,1\}^b$ associated with identifying information (such as user ID, timestamp, or other metadata~\cite{stealthink}), the task is to encode $\mathbf{m}$ into the model’s output to produce a watermarked text $\mathbf{x}_g'$ via an \emph{encoder}: $
\mathbf{x}_g'=\mathtt{Enc}(f,\mathbf{x}_p,\mathbf{m}),
$
while preserving fluency and semantics relative to the unwatermarked output $\mathbf{x}_g$. 
% For every user prompt, the provider derives a binary message that encodes identifying information (e.g., user ID, generation timestamp, or other metadata)~\cite{stealthink}, and embeds it into the output via a message \emph{encoder}. 
When a suspicious text is later encountered, a \emph{decoder} can be applied to recover the encoded message: $
\mathbf{m}'=\mathtt{Dec}(\mathbf{x}_g')
$, enabling provenance verification and attribution of misuse to the originating account. %~\cite{liu2024survey, wu2025survey, pan2024markllm}.
% 
% %We formalize multi-bit watermarking as follows. 
% Given a LLM $f$, a user prompt $\mathbf{x}_p$, and a $b$\footnote{Throughout, we assume $b$ is odd.}-bit binary message $\mathbf{m}\in\{0,1\}^b$, the multi-bit watermarking task is to first encode $\mathbf{m}$ into the model’s output to produce a watermarked text $\mathbf{x}_g'$ via  
% % An encoder $\mathtt{Enc}(\cdot)$ performs this operation,
% $
% \mathbf{x}_g'=\mathtt{Enc}(f,\mathbf{x}_p,\mathbf{m}),
% $
% while preserving fluency and semantics relative to the unwatermarked output $\mathbf{x}_g$. 
% Then, the encoded message can be decoded via 
% %A decoder $\mathtt{Dec}(\cdot)$ is then applied to recover the message,
% $
% \mathbf{m}'=\mathtt{Dec}(\mathbf{x}_g').
% $
The objective of multi-bit watermarking is to maximize the probability of exact message recovery subject to a minimum text quality constraint:
\begin{align*}
    \max \quad & \Pr\!\Big[\mathbf{m}'=\mathbf{m}\ \big|\ \mathbf{m}'=\mathtt{Dec}(\mathtt{Enc}(f,\mathbf{x}_p,\mathbf{m}))\Big] \\
    \text{s.t.} \quad & \mathtt{Quality}(\mathbf{x}_g'\mid \mathbf{x}_p)\ \ge\ \tau,
\end{align*}
where $\mathtt{Quality}(\cdot)$ measures text quality and $\tau$ is a threshold calibrated to approximate the quality of $\mathbf{x}_g$. 
In practice, conditional perplexity $\mathtt{PPL}(\cdot)$ is a common choice for $\mathtt{Quality}(\cdot)$. 
This formulation ensures highly reliable message recovery with negligible degradation in text quality.

\begin{figure}
    \centering
    \includegraphics[width=\linewidth]{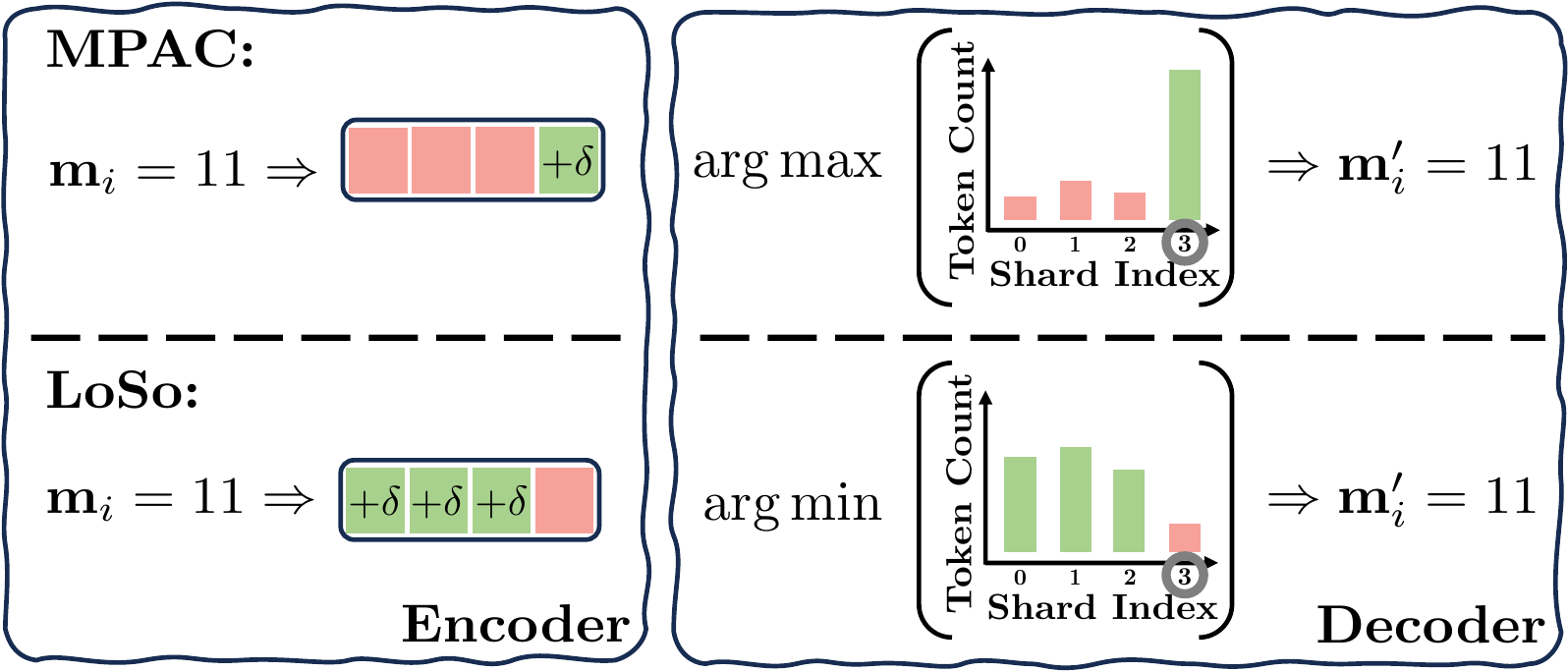}
    \caption{
    Illustration of the encoder and decoder for MPAC (top) and \textsc{LoSo} (bottom).
    }
    \label{fig: mpac}
\end{figure}

\subsection{Classic Solution}

% \textbf{Revisiting MPAC.} In this section, we revisit 
To solve the above problem, the encoder of the classic multi-bit watermarking method MPAC~\cite{mpac} divides the message $\mathbf{m}$ into $r$ blocks $\{\mathbf{m}_0, \mathbf{m}_1, \dots, \mathbf{m}_{r-1}\}$, each containing $d\geq2$ bits. 

\textbf{Encoder:} 
%
% Then, during each token generation step of the model $f$, one message block is pseudo-randomly selected for encoding. 
%
For generating an output sequence of $T$ tokens, in the $t$-th ($t \in [T]$) token generation step of the model $f$, the encoding process of MPAC includes the following three steps:

\ding{172} Logits generation: The model computes the logits vector $\ell_{t} = [\ell_{t}^v]_{v \in \mathcal{V}} 
%= \bigl[f(v \mid \mathbf{x}_p, \mathbf{x}_{:t-1})\bigr]_{v \in \mathcal{V}}
,$ where $\ell_{t}^v = f(v \mid \mathbf{x}_p, \mathbf{x}_{:t-1})$ denotes the logit score assigned to token $v$ in its vocabulary $\mathcal{V}$ of size $V$ and  $\mathbf{x}_{:t-1} = \{x_1, x_2, \ldots, x_{t-1}\}$ denotes the generated tokens. 
% 
% To embed a block $\mathbf{m}_i, \, i \in [r-1]$, the encoder first derive a hash seed $\mathbf{s}$ from a hash function $\mathtt{Hash}(\cdot)$ with a hash key $\mathbf{k}$. Then it use $\mathbf{s}$ to permute the vocabulary $\mathcal{V}$, yield $\mathcal{V}^\prime$ and partitions $\mathcal{V}^\prime$ into $2^d$ disjoint shards $\{\mathcal{S}_0, \mathcal{S}_1, \dots, \mathcal{S}_{2^d - 1} \}$. 
% 

\ding{173} Logits perturbation: A message block $\mathbf{m}_i$ is
pseudo-randomly selected (e.g., $i = x_{t-1} \mod r$). 
% pick a message block $\mathbf{m}_i$ (e.g., let $i = x_{t-1} \mod r$). 
The encoder then computes a hash seed $\mathbf{s}=\mathtt{Hash}(x_{t-1} , \mathbf{k})$, where $\mathtt{Hash}(\cdot)$ is a hash function and $\mathbf{k}$ is a pre-defined hash key. This seed is used to permute the vocabulary, producing a permuted vocabulary $\mathcal{V}'$. Next, the encoder partitions $\mathcal{V}'$ 
into $2^d$ disjoint shards $\{\mathcal{S}_0,\mathcal{S}_1,\ldots,\mathcal{S}_{2^d-1}\}$, and designates the $[\mathbf{m}_i]_{10}$-th shard as the green list $\mathcal{G}$, where $[\cdot]_{10}$ denotes the decimal value of the binary message block. This results in a green list ratio of $\gamma= |\mathcal{G}|/ |\mathcal{V}|=2^{-d}$. Finally, the logits of tokens in $\mathcal{G}$ are boosted by adding a watermarking bias $\delta>0$, yielding perturbed logits $\ell^\prime_t$. 

A concrete example is shown in \autoref{fig: mpac}. 
Suppose the block length is $d=2$ and the selected message block is $\mathbf{m}_i = 11$. 
The permuted vocabulary $\mathcal{V}^\prime$ is then divided into $2^d = 4$ shards $\{\mathcal{S}_0, \mathcal{S}_1, \mathcal{S}_2, \mathcal{S}_3 \}$, and since $[11]_{10} = 3$,  $\mathcal{S}_3$ is selected as $\mathcal{G}$, and its logits are perturbed. 

\ding{174} Token sampling: With the perturbed logits $\ell_t^\prime$, the model completes the token generation by converting $\ell_t^\prime$ into a probability distribution over $\mathcal{V}$ via $\mathtt{softmax}(\cdot)$ and sampling a token $x_t^\prime$ from it. The encoder repeats these three steps for each generation step and finally obtains the watermarked sequence $\mathbf{x}_g'=\{x_t^\prime\}^{T}_{t=1}$.

\textbf{Decoder: } Given a suspect text sequence $\mathbf{x}_s$ consisting of $T$ tokens, the MPAC decoder examines each token $x_t^\prime \in \mathbf{x}_s$. For each token, it first determines the index $i$ of the message block that the token corresponds to, and reconstructs the shard partitions using the hash key $\mathbf{k}$. It then updates a token-shard mapping matrix (TMM) $\mathbf{A} \in \mathbb{N}^{r \times 2^d}$ by incrementing $\mathbf{A}[i, u]$ whenever $x_t^\prime$ belongs to the shard $\mathcal{S}_u$.
% the decoder enumerates all generated tokens. It 
% the decoder first initializes a token frequency matrix $\mathbf{A}\in \mathbb{N}^{r\times 2^d}$. For each token, it identifies the index $i$ of the message block carried by the token and records the shard $u$ to which the token belongs (incrementing $\mathbf{A}[i, u]$ by $1$). 
% 
After processing all tokens, the decoder recovers the decimal value of the $i$-th message block $p^\prime_i$ as
\[
p_i^\prime \;=\; \arg\max_{u\in\{0,\ldots,2^d-1\}} \mathbf{A}[i,u],
\]
that is, the index of the shard with the highest token-shard mapping counts. This shard corresponds to $\mathcal{G}$, since $\mathcal{G} = \mathcal{S}_{[\mathbf{m}_i]_{10}}$, and logits of tokens in $\mathcal{G}$ were boosted during encoding, causing them to appear more frequently in the suspect text (see Figure~\ref{fig: mpac}). 
% since the shard carrying the watermark signal has been perturbed and thus exhibits a higher token frequency (see Figure~\ref{fig: mpac}).
% 
Finally, the block message is recovered by converting $p_i^\prime$ into its $d$-bit binary representation. 
%
% The decoded values are then converted into their $d$-bit binary representations, and the complete message is reconstructed by concatenating all decoded blocks.

\begin{figure}
    \centering
    \includegraphics[width=\linewidth]{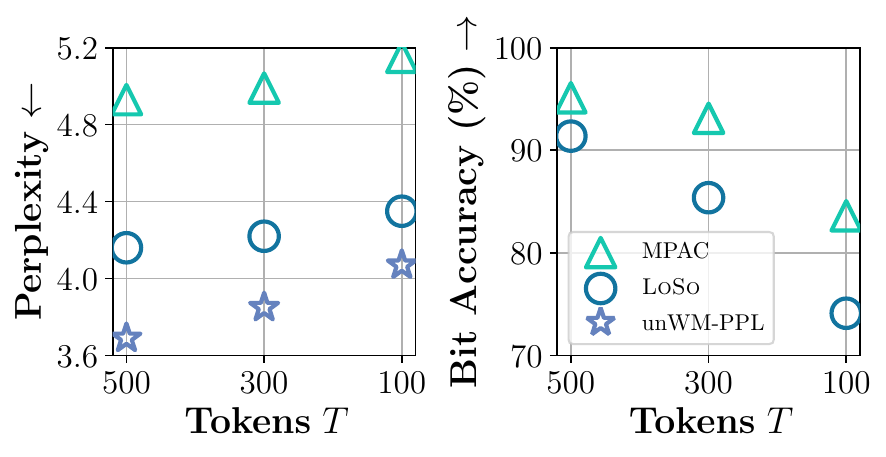}
    \caption{Left: Perplexity vs.\ $T$ for MPAC, \textsc{LoSo}, and clean (unwatermarked) text. Right: Bit accuracy vs.\ $T$ for MPAC and \textsc{LoSo}. Note that we do not report bit accuracy for clean text, since it corresponds to the false-positive case. Results are from the text completion task on LLaMA-$2$-$7$B~\cite{llama2} using C$4$ dataset~\cite{c4}, with message length $b=16$ under varying output sequence lengths $T$.}
    \label{fig: intuition}
\end{figure}

Empirically, MPAC achieves good decoding accuracy when the text sequence is sufficiently long for decoding, as more tokens lead to more accurate token-shard mapping estimates, thereby improving the reliability of correctly identifying the green list. 
As shown in Figure~\ref{fig: intuition}, when $T=500$, MPAC reaches a bit accuracy (i.e., the proportion of bits correctly decoded from the text) of $95.12\%$. However, when $T$ decreases to $100$, the bit accuracy drops significantly to $83.62\%$. Moreover, another major limitation of MPAC is that it largely degrades the quality of the generated texts: MPAC significantly increases the perplexity compared to the unwatermarked text. 
% As noted in prior works, 
In fact, a larger green list ratio better preserves the quality of watermarked text, since it induces less distortion in the overall logits distribution~\cite{rsbh, majormark}. However, MPAC is constrained to $\gamma \leq 0.25$ for any $d \geq 2$ by design, leading to a sharp distortion. %This effect is also shown in Figure~\ref{fig: intuition}, where MPAC significantly increases the perplexity compared to unwatermarked text. 

\section{Our Method: \textsc{XMark}}

We propose \textsc{XMark}, a novel multi-bit watermarking method that incorporates three key designs (\textsc{LoSo}, evergreen list, and constrained TMM) to improve text quality and decoding accuracy.

\IncMargin{0.2em}
\begin{algorithm}[t]
\normalsize
\caption{The Encoder of \textsc{XMark}}
\label{alg: XMark_encode}
\SetKwInOut{Input}{Input}
\SetKwInOut{Output}{Output}
\Input{User prompt $\mathbf{x}_p$, LLM $f$, bias $\delta$, vocabulary $\mathcal{V}$, maximum length $T$, message $\mathbf{m} \in \{0,1\}^b$, number of blocks $r$, hash keys $\{\mathbf{k}_j\}_{j=0}^{k-1}$.}
\Output{Watermarked text $\mathbf{x}_g^\prime$.}
% $\mathbf{x}_{:2} \leftarrow \mathbf{x}_{p, -1:-2}$

$\mathbf{x}_{:2} \leftarrow \{x_{p, -2}, x_{p, -1} \}$

% $\mathbf{x}^\prime_g\leftarrow \{x_{p, -2}, x_{p,-1}\}$

% Divide $\mathbf{m}$ into $r$ blocks: 
$\mathbf{m}_0, \dots, \mathbf{m}_{r-1} \leftarrow \mathtt{divide}(\mathbf{m}, r)$ \label{algline: message_division}

$d \leftarrow b/r$ \label{algline: message_block_length}

\For{$t \leftarrow 1$ \KwTo $T$}{
    $\ell_{t} \leftarrow f(\mathbf{x}_p, \mathbf{x}_{3:})$ \label{algline: get_logits}

    % $x_{t-2}, x_{t-1}  \leftarrow \mathbf{x}_{t-2}, \mathbf{x}_{t-1}$ \label{algline: previous_two_tokens}
    
    $i \leftarrow (x_{t-2} + x_{t-1}) \bmod r$ \label{algline: get_message_block}

    $p \leftarrow [\mathbf{m}_i]_{10}$ \label{algline: p_calculation}
    
    \For{$j \leftarrow 0$ \KwTo $k-1$}{
        $\mathbf{s}_j \leftarrow \mathtt{Hash}(x_{t-2}, x_{t-1} , \mathbf{k}_j)$ \label{algline: get_hash_key}

        $\mathcal{V}^\prime_j \leftarrow \mathtt{permute}(\mathcal{V}, \mathbf{s}_j)$ \label{algline: permutation_over_vocab}

        $\mathcal{S}_{j,0}, \dots, \mathcal{S}_{j,2^d-1} \leftarrow \mathtt{part}(\mathcal{V}^\prime_j, 2^d)$ \label{algline: shard_partition}

        $\mathcal{G}_j \leftarrow \mathcal{V}^\prime_j \setminus \mathcal{S}_{j, p} $ \label{algline: green_list_construct}
    }

    $\mathcal{E} \leftarrow \bigcap^{k-1}_{j=0}\mathcal{G}_j$ \label{algline: construct_evergreen}
    
    $\ell^{\prime}_t \leftarrow \ell_t $
    
    \For {$v \in \mathcal{E}$}{\label{algline: perturb_start}
        $\ell^{\prime, v}_{t} \leftarrow \ell^{ \prime,v}_{t} + \delta$
        } \label{algline: perturb_end}

    Sample $x_t^\prime \sim \mathtt{Softmax}(\ell^{\prime}_{t})$  \label{algline: sample_token}
    
    Append $x_t^\prime$ to $\mathbf{x}$
    
}
\textbf{Return} $\mathbf{x}^\prime_{g}=\mathbf{x}_{3:}$ \\
\end{algorithm}

\subsection{Leave-one-shard-out Watermarking}
Recall that in MPAC, the vocabulary shard indexed by the block value $[\mathbf{m}_i]_{10}$ is selected as the green list. In contrast, we propose a method called Leave-one-Shard-out (\textsc{LoSo}) watermarking, which reverses this choice by marking \emph{all} shards \emph{except} the $[\mathbf{m}_i]_{10}$-th as green (leaving $[\mathbf{m}_i]_{10}$-th shard unperturbed). 
For example (see Figure~\ref{fig: mpac}), when $d=2$ and $\mathbf{m}_i=11$, $\mathcal{S}_0$, $\mathcal{S}_1$, and $\mathcal{S}_2$ form the green list while $\mathcal{S}_3$ is excluded. 
During decoding, the block value is inferred as the index of the shard with \emph{fewest} token-shard mapping counts, i.e., the excluded one.
This design achieves a larger green list ratio of $\gamma = 1 - 2^{-d} \ge 0.75$ for any $d \ge 2$, compared to $\gamma = 2^{-d}$ in MPAC, thereby substantially improving text quality. As shown in Figure~\ref{fig: intuition}, \textsc{LoSo} yields significantly lower perplexity than MPAC and is closer to the quality of unwatermarked text. However, this gain in text quality comes with a cost of reduced decodability, since enlarging the green list weakens the watermark signal. 
%, so reliable recovery requires more tokens. 
As shown in Figure~\ref{fig: intuition}, when $T=500$, \textsc{LoSo} attains $91.38\%$ bit accuracy versus $95.12\%$ for MPAC (a $3.74\%$ gap). This gap goes up to $9.50\%$ when $T=100$. 
% with $T=100$ tokens \textsc{LoSo} attains $74.12\%$ bit accuracy versus $83.62\%$ for MPAC (a $9.5\%$ gap); at $T=500$, the gap narrows to $3.74\%$. 

% In practice, collecting sufficient tokens is often challenging; decoding accuracy degrades in limited-token settings, and the effect is amplified as the message length $b$ increases.

In the following, based on \textsc{LoSo}, we propose a novel method, named \textsc{XMark}, which preserves the quality of watermarked text while ensuring high decoding accuracy, even when the number of tokens available for decoding is limited. We provide the detailed algorithm of the encoder of \textsc{XMark} in Algorithm~\ref{alg: XMark_encode}. The algorithm of the decoder can be found in the Appendix~\ref{apdx: decoder_alg}.

\subsection{Encoder of \textsc{XMark}}

We first describe the encoding process of \textsc{XMark}, and then analyze its advantages in preserving text quality. Specifically, before text generation starts, \textsc{XMark} first divides the binary message $\mathbf{m}$ into $r$ blocks $\{\mathbf{m}_{0}, \mathbf{m}_1, \dots, \mathbf{m}_{r-1}\}$, where each block has $d\geq2$ bits (Line~\ref{algline: message_division}-\ref{algline: message_block_length}). It also prepares $k$ hash keys $\{\mathbf{k}_0, \mathbf{k}_1, \dots, \mathbf{k}_{k-1}\}$.  
% Note that in XMark, it fixes each message block consisting of $d=2$ bits (i.e., $r = b/2$).

At the $t$-th generation step for token $x_t$, the encoder of \textsc{XMark} obtains the model logits $\ell_t$ based on the user prompt and previous generated tokens (Line~\ref{algline: get_logits}). It samples a message block $\mathbf{m}_i$ to be embedded, where 
$i = (x_{t-2} + x_{t-1}) \bmod r$ (Line~\ref{algline: get_message_block}). Given a hash function 
% $\mathtt{Hash}(x_{t-2}, x_{t-1} \mid \cdot)$ 
that takes as input the two previously generated token IDs and a hash key, it obtains $k$ distinct hash seeds $\{\mathbf{s}_j\}_{j=0}^{k-1}$, i.e., $\mathbf{s}_j=\mathtt{Hash}(x_{t-2},x_{t-1}, \mathbf{k}_j), \forall j \in \{0, 1, \dots, k-1 \}$ %corresponding to the $k$ keys 
(Line~\ref{algline: get_hash_key}).  
Using these seeds, the encoder permutes the vocabulary $\mathcal{V}$ to generate $k$ different permutations: $\{\mathcal{V}_{j}^\prime\}_{j=0}^{k-1}$ (Line~\ref{algline: permutation_over_vocab}). Each permuted vocabulary $\mathcal{V}_j^\prime$ is then evenly partitioned into $2^d$ shards: $\mathcal{V}_j^\prime = \{\mathcal{S}_{j,0}, \mathcal{S}_{j,1}, \dots, \mathcal{S}_{j,2^d-1}\}$ (Line~\ref{algline: shard_partition}). For $\mathcal{V}_j^\prime$, the corresponding green list is constructed via \textsc{LoSo} as $\mathcal{G}_j = \mathcal{V}_j^\prime \setminus \mathcal{S}_{j,p}$ (Line~\ref{algline: green_list_construct}), where $p = [\mathbf{m}_i]_{10}$ (Line~\ref{algline: p_calculation}). The intersection of all $k$ green lists constructs an \textit{evergreen list} $\mathcal{E} = \bigcap_{j=0}^{k-1}\mathcal{G}_j$ (Line~\ref{algline: construct_evergreen}).
Finally, the logits of all tokens in $\mathcal{E}$ are perturbed by a watermarking bias $\delta>0$, and the token $x_t^\prime$ is sampled correspondingly (Line~\ref{algline: perturb_start}-\ref{algline: sample_token}). This process continues until all $T$ tokens are generated. 

% is added to the logits of all tokens in $\mathcal{E}$ (Line~\ref{algline: perturb_start}-Line~\ref{algline: perturb_end}). A softmax function is then applied to the perturbed logits to produce a sampling probability over vocabulary, and the $x_t$ is correspondingly sampled (Line~\ref{algline: sample_token}). This process will continue until all $T$ tokens are generated. 

\begin{figure}
    \centering
    \includegraphics[width=\linewidth]{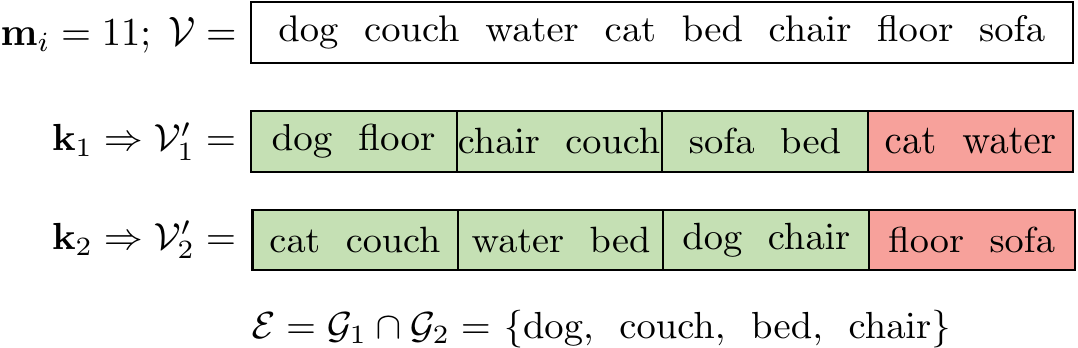}
    \caption{Illustration showing that tokens in $\mathcal{E}$ may originate from different shards across multiple permutations.
    }
    \label{fig: motivation}
\end{figure}

By using the \textsc{LoSo} strategy, the encoder of \textsc{XMark} achieves an improved green list ratio, thereby preserving the quality of the watermarked texts. The following theorem characterizes the expected green list ratio of \textsc{XMark}. 
% (i.e., the length of $\mathcal{E}$) 
% under $k$ hash keys.

\begin{theorem}[$\gamma$ of \textsc{XMark}]\label{them:famark_gls}
The evergreen list $\mathcal{E}$ constructed by \textnormal{\textsc{XMark}} satisfies $\mathbb{E}[|\mathcal{E}|] \approx (1-2^{-d})^k |\mathcal{V}|$, i.e., $\mathbb{E}[\gamma] =  \mathbb{E}[|\mathcal{E}|/|\mathcal{V}| ]\approx (1-2^{-d})^k.$
\begin{proof}
    The proof is in the Appendix~\ref{apdx: proof}.
\end{proof}
\begin{remark}\label{remark: them}
(1) Discussion on $d$: For a fixed $k$, \textnormal{\textsc{XMark}} produces an exponentially increasing $\gamma$ with $d$, thereby improving text quality. In contrast, if \textnormal{\textsc{XMark}} adopts MPAC’s encoding strategy instead of \textnormal{\textsc{LoSo}}, the expected $\gamma$ becomes $\mathbb{E}[\gamma] = (2^{-d})^k$, which decreases exponentially with $d$ and thus severely degrades text quality. In practice, a large $d$ is not preferred, as it increases the number of candidate shards during decoding and thus requires substantially more tokens to estimate token-shard mapping reliably. In this work, we set $d = 2$ for \textnormal{\textsc{XMark}}.
(2) Discussion on $k$: When $k = 1$, the encoder of \textnormal{\textsc{XMark}} reduces to that of \textnormal{\textsc{LoSo}} (i.e., $\mathcal{E} = \mathcal{G}_0$). When $k \geq 2$, the $\gamma$ of \textnormal{\textsc{XMark}} gradually decreases. However, the evergreen list design can improve decoding accuracy 
(see Section~\ref{sec: decoder_xmark}); thus, $k$ serves as a hyperparameter that balances the trade-off between text quality and decoding accuracy.
\end{remark}
\end{theorem}

\subsection{Decoder of \textsc{XMark}}\label{sec: decoder_xmark}

% Specifically, it initializes a shard-count matrix $\mathbf{A}\in\mathbb{N}^{r\times 2^d}$ with zeros.
Here, we first present the details of the decoding process of \textsc{XMark} and then discuss the importance of the evergreen list and constrained TMM for decoding accuracy.  

Given a suspect text $\mathbf{x}_s$, the decoder first initializes a \textit{constrained token-shard mapping matrix} (cTMM) $\mathbf{A}\in \mathbb{N}^{r\times 2^d}$ as all zeros, i.e., $\mathbf{A}^0=\mathbf{0}$. For each token $x_t^\prime \in \mathbf{x}_s$,
% with $t\in\{3, 4, \dots, T\}$, \footnote{We discard $x_1$ and $x_2$ because computing the block index and hash seed requires $x_{t-2}$ and $x_{t-1}$, which depend on the (generally unavailable) prompt tokens.}
the decoder of \textsc{XMark} first computes current message block index $i$ as $i= (x_{t-2}^\prime+x_{t-1}^\prime) \bmod r$. Given each hash key used in the encoder $\mathbf{k}_j, \forall j \in \{0, 1, \dots, k-1 \}$, it reconstructs each permuted vocabulary $\mathcal{V}_j^\prime$ with the seed $\mathbf{s}_j=\mathtt{Hash}(x_{t-2}^\prime,x_{t-1}^\prime, \mathbf{k}_j)$ and partitions $\mathcal{V}_j^\prime$ into $2^d$ shards $\{\mathcal{S}_{j,u}\}_{u=0}^{2^d-1}$. If $x_t^\prime\in \mathcal{S}_{j,u}$, the decoder 
updates the cTMM $\mathbf{A}^t \in\mathbb{N}^{r\times 2^d}$ by incrementing $ \mathbf{A}[i,u]$ by 1, subject to the \textit{constraint} $\mathbf{A}^t[i,:] - \mathbf{A}^{t-1}[i, :] \in \{0,1\}^{2^d}$. This process continues until all tokens in $\mathbf{x}_s$ are enumerated. 
% We refer to this method as \textit{greedy contribution counting} (\textsc{Gcc}).

For each block $i\in\{0,\ldots,r-1\}$, the decoder determines the decimal value of $i$-th block $p_i^\prime$ by
\[
  p_i^\prime \;=\; \arg\min_{u\in\{0,\ldots,2^d-1\}} \mathbf{A}[i,u],
\]
since the shard with the \textit{fewest} token–mapping counts is more likely unboosted during encoding.
% \xu{since the shard with the least generated tokens is more likely to remain not boosted during encoding.}
Finally, it converts each $p_i^\prime$ to its $d$-bit binary representation and concatenates them to recover the full message.

\textbf{Evergreen List vs \textsc{LoSo} Green List.} 
Recall the decoding challenge of existing methods when given a short suspect text sequence, which cannot provide sufficient tokens for extracting the message. 
The evergreen list in \textsc{XMark} provides more observations for the token-shard mapping. Specifically, when $k=1$, the evergreen list reduces to the \textsc{LoSo}'s green list, $\sum\mathbf{A}=T$ indicates exactly $T$ observations that construct the mapping between the token and its originating shard. When $k \geq 2$, the number of observations increases to at most $kT$. 
An example is given in Figure~\ref{fig: motivation}. With $k=2$ and $\mathbf{m}_i=11$, the token “\textit{dog}” in $\mathcal{E}$ belongs to $\mathcal{S}_{1,0}$ of $\mathcal{V}_1^\prime$ and $\mathcal{S}_{2,2}$ of $\mathcal{V}_2^\prime$. 
% Therefore, after reconstructing all $k$ permutations during decoding, each observed token contributes up to $k$ votes, at most one per permutation, to the \hu{shard counts} for each message block. The resulting increase in effective counts makes the perturbed shard more likely to emerge as the \textit{argmax}, thereby improving decoding accuracy, especially when the number of available tokens is limited.
% Therefore, after reconstructing all $k$ permutations during decoding, each observed token contributes up to $k$ updates to cTMM. 
This mitigates the problem of inaccurate construction of token-shard mapping due to the limited tokens. 

% The resulting increase in effective frequency makes the perturbed shard more likely to be correctly identified, thereby addressing the challenges of the limited tokens for decoding.

\textbf{cTMM vs TMM.} In existing methods and also our naive \textsc{LoSo} method, a TMM is used for decoding. 
While intuitive, the TMM results in each token–shard mapping being counted up to 
$k$ times in the extreme case, when the evergreen list is used during encoding.  
For instance, consider a generated token belonging to $\mathcal{V}\setminus \bigcup_{j=0}^{k-1}\mathcal{G}_j$, 
i.e., it does not appear in any green list across all permutations. 
In this case, the token would be counted $k$ times into the same unboosted shard  (like the shard $3$ of \textsc{LoSo} in Figure~\ref{fig: mpac}), making it difficult to distinguish between the boosted shards and the unboosted shard, especially when the limited number of tokens does not provide sufficient observations for reliable token-shard mapping construction. 
cTMM addresses this issue by constraining each token to contribute at most once to any shard, effectively preventing the explosion of counts for the unboosted shard. 

\begin{table*}[ht]
  \centering
  \caption{
Comparison of methods on three downstream tasks with $b=8$ under different token budgets $T \in \{150, 200, 250, 300\}$. The average PPL of unwatermarked texts is $3.97$, $4.59$, and $3.69$ for the three tasks, respectively, serving as the lower bounds for the PPLs of watermarked texts.
}
  \scalebox{0.8}{
    \begin{tabular}{lcccccccc|cc}
    \toprule
    % \footnotesize
    % \multirow{2}[2]{*}{\textbf{\makecell*[c]{Task \\ (Dataset)}}} 
     \multirow{2}[2]{*}{\textbf{Method}} 
    & \multicolumn{2}{c}{$T=150$}   
    & \multicolumn{2}{c}{$T=200$}   
    & \multicolumn{2}{c}{$T=250$} 
    & \multicolumn{2}{c|}{$T=300$} 
    & \multirow{2}[2]{*}{\textbf{\makecell*[c]{Avg. \\ BA$\uparrow$}}} 
    & \multirow{2}[2]{*}{\textbf{\makecell*[c]{Avg. \\ PPL$\downarrow$}}} \\
    \cmidrule(r){2-3}\cmidrule(r){4-5}\cmidrule(r){6-7}\cmidrule(r){8-9}
             & \textbf{BA}$\uparrow$    & \textbf{PPL}$\downarrow$   
          & \textbf{BA}$\uparrow$    & \textbf{PPL}$\downarrow$   
          & \textbf{BA}$\uparrow$    & \textbf{PPL}$\downarrow$
          & \textbf{BA}$\uparrow$    & \textbf{PPL}$\downarrow$ \\
    \midrule
    \textit{Text Completion} \\
     CycleShift      & $95.25$ & $5.17$ & $98.00$ & $5.02$ & $97.00$ & $5.07$ & $98.25$ & $4.96$ & $97.13$ & $5.06$ \\
     DepthW          & $79.50$ & $4.62$ & $89.00$ & $4.67$ & $90.50$ & $4.63$ & $95.75$ & $4.62$ & $88.69$ & $4.64$ \\
     StealthInk       & $85.00$ & $4.13$ & $89.50$ & $4.21$ & $92.25$ & $4.13$ & $92.50$ & $4.04$ & $89.81$ & $4.13$ \\
     MPAC            & $94.00$ & $5.19$ & $94.00$ & $5.00$ & $96.25$ & $5.03$ & $98.25$ & $5.10$ & $95.63$ & $5.08$ \\
     RSBH            & $92.75$ & $4.83$ & $96.00$ & $4.83$ & $95.50$ & $4.67$ & $97.75$ & $4.83$ & $95.50$ & $4.79$ \\
     \cellcolor[rgb]{ .9, .9, .9}\textsc{XMark} & \cellcolor[rgb]{ .9, .9, .9}$98.75$ & \cellcolor[rgb]{ .9, .9, .9}$4.68$ & \cellcolor[rgb]{ .9, .9, .9}$99.00$ & \cellcolor[rgb]{ .9, .9, .9}$4.61$ & \cellcolor[rgb]{ .9, .9, .9}$99.75$ & \cellcolor[rgb]{ .9, .9, .9}$4.60$ & \cellcolor[rgb]{ .9, .9, .9}$100.00$ & \cellcolor[rgb]{ .9, .9, .9}$4.56$ & \cellcolor[rgb]{ .9, .9, .9}$99.38$ & \cellcolor[rgb]{ .9, .9, .9}$4.61$ \\
    \midrule \midrule
    % \multirow{6}[2]{*}{\makecell*[c]{Text Summarization \\ (CNN-DailyNews)}} 
    \textit{Text Summarization} \\
    CycleShift      & $54.50$ & $5.57$ & $59.25$ & $4.82$ & $64.75$ & $4.36$ & $62.25$ & $4.10$ & $60.19$ & $4.71$ \\
     DepthW          & $58.50$ & $5.60$ & $55.00$ & $4.76$ & $57.25$ & $4.29$ & $54.00$ & $3.95$ & $56.19$ & $4.65$ \\
     StealthInk       & $62.75$ & $5.55$ & $65.00$ & $4.60$ & $69.50$ & $4.26$ & $71.50$ & $3.93$ & $67.19$ & $4.59$ \\
     MPAC            & $71.75$ & $5.71$ & $75.75$ & $4.94$ & $78.25$ & $4.36$ & $82.00$ & $4.13$ & $76.94$ & $4.79$ \\
     RSBH            & $62.50$ & $5.60$ & $65.00$ & $5.04$ & $63.75$ & $4.56$ & $64.75$ & $4.13$ & $64.00$ & $4.83$ \\
     \cellcolor[rgb]{ .9, .9, .9}\textsc{XMark} & \cellcolor[rgb]{ .9, .9, .9}$74.75$ & \cellcolor[rgb]{ .9, .9, .9}$5.50$ & \cellcolor[rgb]{ .9, .9, .9}$79.00$ & \cellcolor[rgb]{ .9, .9, .9}$4.85$ & \cellcolor[rgb]{ .9, .9, .9}$87.75$ & \cellcolor[rgb]{ .9, .9, .9}$4.33$ & \cellcolor[rgb]{ .9, .9, .9}$89.50$ & \cellcolor[rgb]{ .9, .9, .9}$3.97$ & \cellcolor[rgb]{ .9, .9, .9}$82.75$ & \cellcolor[rgb]{ .9, .9, .9}$4.66$ \\
    \midrule \midrule 
    \textit{Story Generation} \\
     CycleShift      & $66.00$ & $4.93$ & $66.00$ & $4.21$ & $66.75$ & $3.80$ & $74.00$ & $3.58$ & $68.19$ & $4.13$ \\
     DepthW          & $56.25$ & $4.78$ & $59.75$ & $3.99$ & $55.75$ & $3.60$ & $54.75$ & $3.36$ & $56.63$ & $3.93$ \\
     StealthInk       & $59.00$ & $4.55$ & $62.50$ & $3.88$ & $69.00$ & $3.44$ & $71.25$ & $3.22$ & $65.44$ & $3.77$ \\
     MPAC            & $71.00$ & $4.98$ & $71.25$ & $4.24$ & $76.25$ & $3.83$ & $78.00$ & $3.61$ & $74.13$ & $4.17$ \\
     RSBH            & $58.00$ & $4.96$ & $62.75$ & $4.33$ & $65.75$ & $3.85$ & $77.00$ & $3.47$ & $65.88$ & $4.15$ \\
     \cellcolor[rgb]{ .9, .9, .9}\textsc{XMark} & \cellcolor[rgb]{ .9, .9, .9}$75.50$ & \cellcolor[rgb]{ .9, .9, .9}$4.86$ & \cellcolor[rgb]{ .9, .9, .9}$79.75$ & \cellcolor[rgb]{ .9, .9, .9}$4.12$ & \cellcolor[rgb]{ .9, .9, .9}$82.50$ & \cellcolor[rgb]{ .9, .9, .9}$3.68$ & \cellcolor[rgb]{ .9, .9, .9}$86.00$ & \cellcolor[rgb]{ .9, .9, .9}$3.44$ & \cellcolor[rgb]{ .9, .9, .9}$80.94$ & \cellcolor[rgb]{ .9, .9, .9}$4.03$ \\
    \bottomrule
    \end{tabular}%
    }
  \label{tab:main_result_table}%
\end{table*}

\section{Empirical Evaluations}

\subsection{Experimental Settings}
\textbf{General Settings.}
By default, we consider a total of $50$ users, corresponding to $50$ randomly generated messages to be encoded. Each user submits two prompts, and each prompt is used to generate a text of $T/2$ tokens, resulting in $T$ tokens per user for decoding. Unless otherwise specified, we set $T=150$ by default. We compare our proposed methods with five state-of-the-art methods, including DepthW~\cite{depthw}, CycleShift~\cite{cycleshift}, MPAC~\cite{mpac}, RSBH~\cite{rsbh}, and StealthInk~\cite{stealthink}. For DepthW and CycleShift, we only evaluate the case of $b = 8$, as their decoding time grows exponentially and becomes impractical for larger message lengths. For \textsc{XMark}, we set $d=2$ and $k=2$ by default. These settings result in an expected green list ratio of $\mathbb{E}[\gamma]\approx0.5625$, which is larger than those of existing methods ($0.25$ for MPAC and CycleShift and $0.5$ for DepthW and RSBH). We set the watermarking bias to $\delta=2$ for all methods by default, except for StealthInk, which does not require $\delta$.

\textbf{Datasets and Models.} 
We evaluate our method on three widely studied LLM downstream tasks: 
\textit{text completion} on C4 news~\cite{c4}, 
\textit{story generation} on WritingPrompts~\cite{writingprompt}, 
and \textit{text summarization} on CNN/DailyMail~\cite{cnndaily}. 
Unless otherwise specified, we adopt the widely-used open-source LLaMA-$2$-$7$B~\cite{llama2} for the text completion task, 
and the LLaMA-$2$-$7$B-chat for the story generation and text summarization tasks. 
The system prompts used for these tasks are provided in the Appendix~\ref{apdx: systemprompt}.

\textbf{Metrics.}
We evaluate the quality of generated text using perplexity (PPL), computed by a larger LLaMA-$2$-$13$B model. In addition, we consider the semantic metric BERTScore (BSc.)~\cite{bertscore} and the lexical metrics ROUGE-1 (R.-1) and ROUGE-Lsum (R.-Lsum)~\cite{lin2004rouge}. Following prior practice, these quality metrics are computed by comparing the watermarked text with the corresponding unwatermarked text generated by the same model. For decoding performance, we adopt bit accuracy (BA), following prior works~\cite{ba1,rsbh,mpac, majormark, stealthink}, which is defined as the proportion of correctly decoded bits with respect to the ground-truth message. A desirable method should achieve a low PPL and a high BA, BSc., R.-1, and R.-Lsum.

\subsection{Results}
\textbf{Main Results.} 
Table~\ref{tab:main_result_table} reports the comprehensive BA and PPL results of representative multi-bit watermarking methods on three tasks with $b=8$ and different values of $T \in \{150, 200, 250, 300 \}$. 
As expected, increasing $T$ generally improves BA across all methods and tasks. 
Notably, our method \textsc{XMark} consistently achieves higher BA than all baselines for every $T$ and task. 
On the text summarization and story generation tasks, \textsc{XMark} reaches average BAs of $82.75\%$ and $80.94\%$, outperforming the second-best method MPAC by $+5.81\%$ and $+6.81\%$, respectively. 
On the text completion task, which is inherently more challenging due to higher token entropy, \textsc{XMark} already achieves a near-perfect BA of $98.75\%$ with only $T=150$ tokens, a $+3.5\%$ improvement over CycleShift. 
These results demonstrate \textsc{XMark}'s robustness in ensuring reliable decoding with limited generated tokens. 
This advantage stems from the co-design of the evergreen list and cTMM that enhances the construction of accurate token-shard mapping during decoding. 

In addition to improving BA, \textsc{XMark} preserves the quality of generated texts at a level comparable to or better than existing methods. 
For the text completion task, \textsc{XMark} achieves an average PPL of $4.61$, outperforming all methods except StealthInk. 
However, the high fluency of StealthInk comes at the cost of significantly lower BA. 
For the text summarization and story generation tasks, \textsc{XMark} attains PPLs of $4.66$ and $4.03$, which are only $0.07$ and $0.34$ higher than the unwatermarked baselines, respectively. 
This strong quality preservation stems from the design of \textsc{LoSo}, which allows \textsc{XMark} to maintain a longer green list than existing methods. 

\begin{figure}[t]
    \centering
    \includegraphics[width=\linewidth]{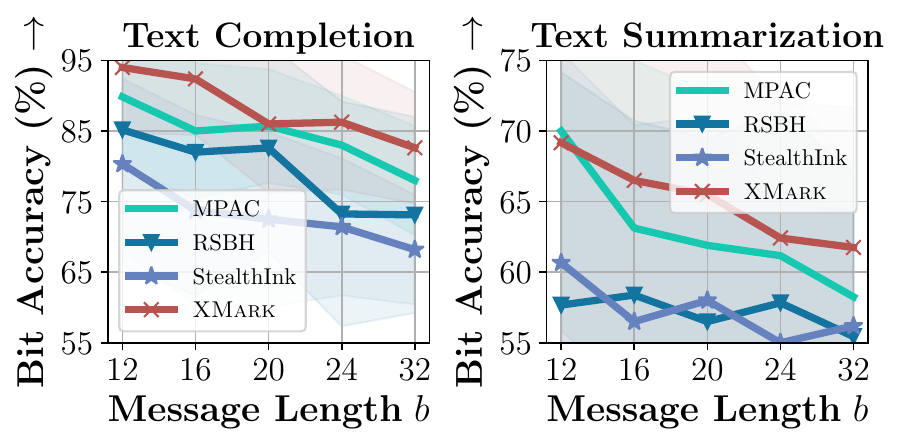}
    \caption{Impact of message length $b$ on BA for MPAC, RSBH, StealthInk, and \textsc{XMark} on the text completion and text summarization tasks. }
    \label{fig: message_length}
\end{figure}
\textbf{Impact of Message Length.} 
In practical applications of watermarking, a binary message length of $b=8$ can encode at most $256$ distinct values (e.g., user IDs), which is often insufficient. 
To evaluate scalability, we extend the message length up to $b=32$ and examine how different methods perform with longer messages. 
Figure~\ref{fig: message_length} reports the BA of MPAC, RSBH, StealthInk, and \textsc{XMark} on the text completion and text summarization tasks. 
As shown, all methods experience a decrease in BA as $b$ increases. 
Nevertheless, \textsc{XMark} consistently outperforms the baselines, maintaining relatively higher BA across different message lengths. 
On the text completion task, even with $b=32$, \textsc{XMark} achieves a BA of $82.62\%$, which is $+4.62\%$ higher than the second-best method, MPAC. 
On the text summarization task, although the absolute BA of all methods becomes low for large $b$, \textsc{XMark} still attains the highest performance, achieving a $+3.5\%$ improvement over MPAC when $b=32$.

\begin{figure}[t]
    \centering
    \includegraphics[width=\linewidth]{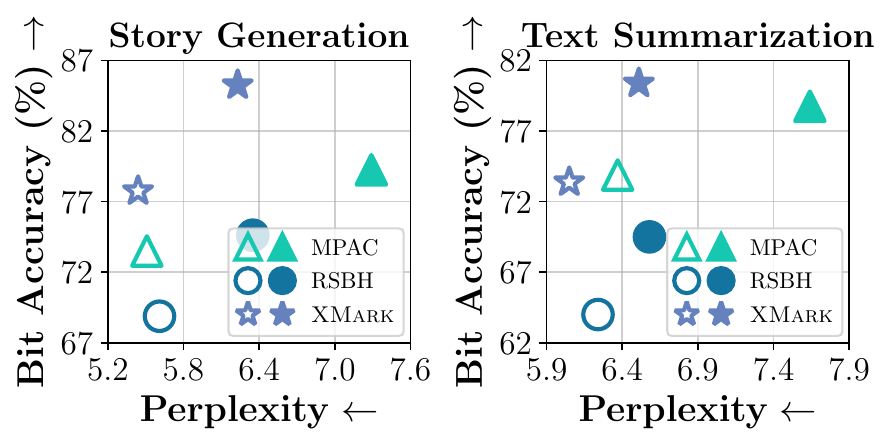}
    % \caption{left sotry gen, right cnn, empty delta= 3, full delta=4}
    \caption{Impact of bias $\delta$ on BA and PPL for MPAC, RSBH, and \textsc{XMark} on the text summarization and story generation tasks. Hollow markers indicate results with $\delta=3$, while solid markers correspond to $\delta=4$.}
    \label{fig: delta}
\end{figure}

\textbf{Impact of Watermarking Bias.} 
We next investigate the impact of different watermarking biases $\delta$ on performance. 
Theoretically, a larger $\delta$ can increase decoding accuracy, but at the cost of degrading text quality since stronger perturbations are applied to the token logits. 
We report the BA and PPL of MPAC, RSBH, and \textsc{XMark} on the text summarization and story generation tasks with $\delta=3$ and $\delta=4$, as shown in Figure~\ref{fig: delta}. 
Overall, \textsc{XMark} consistently achieves a better trade-off between BA and PPL across both settings. 
With $\delta=3$, \textsc{XMark} attains the highest BA while also preserving the lowest PPL on the story generation task. 
When $\delta$ increases to $4$, all methods show improved BA but at the expense of higher PPL, verifying the theoretical analysis of the impact of $\delta$. 
For instance, on the text summarization task, MPAC’s BA increases from $73.88\%$ to $78.55\%$, while its PPL degrades sharply from $6.37$ to $7.64$. 
In contrast, \textsc{XMark} continues to provide the better trade-off, achieving the highest BA ($80.38$) with low PPL ($6.51$) in this setting.

\begin{figure}
    \centering
    \includegraphics[width=\linewidth]{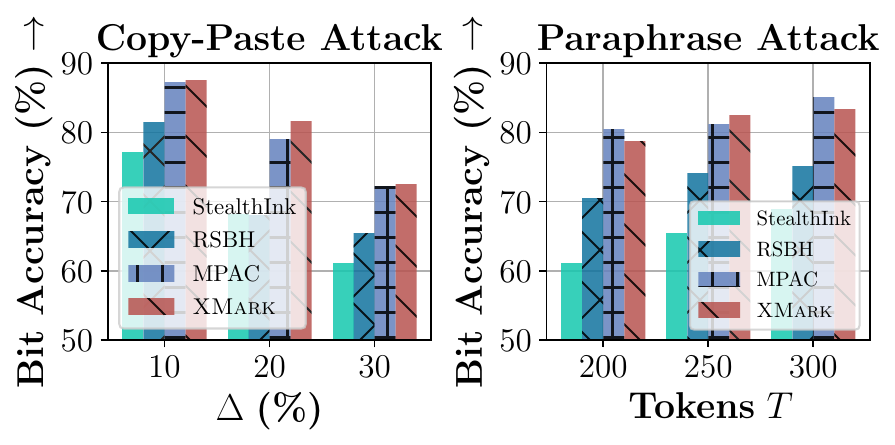}
    \caption{Robustness of methods against text editing attacks on the text completion task with $b=16$.}
    \label{fig: attack}
    % \vspace{-15pt}
\end{figure}

\textbf{Robustness to Text Editing Attacks.} 
When users receive watermarked texts generated by LLMs, they may edit them to improve fluency or, more intentionally, to attempt watermark removal. 
We evaluate the robustness of different methods against two widely studied text editing attacks: \textit{Copy-Paste} and \textit{Paraphrase}~\cite{zhang2024remark}, using the text completion task with $b=16$. 
For the Copy-Paste attack, we mix watermarked and human-written texts by randomly interleaving unwatermarked text segments into watermarked texts, following prior work~\cite{mpac, rsbh}. 
The proportion of unwatermarked text is controlled by $\Delta$, while keeping the total length fixed. 
For the Paraphrase attack, we employ the strong paraphraser \textit{Dipper}~\cite{opengen}. 
We report the BA of StealthInk, RSBH, MPAC, and \textsc{XMark} under varying $\Delta$ for Copy-Paste attack and varying number of generated tokens $T$ under Paraphrase attack in Figure~\ref{fig: attack}.

As shown in Figure~\ref{fig: attack}, for Copy-Paste attack, as $\Delta$ increases, all methods exhibit a decline in BA, as expected. 
Despite this degradation, \textsc{XMark} consistently achieves the highest BA, demonstrating superior robustness. 
Interestingly, MPAC achieves BA comparable to \textsc{XMark}, largely because its design fixes the green list ratio at a small level $0.25$, resulting in a stronger watermark signal that confers additional robustness against editing. 
For Paraphrase attack, which remains a significant open challenge in the literature~\cite{stealthink}, we observe that BA slightly improves for all methods as $T$ increases. 
Here, MPAC performs better than StealthInk and RSBH, while \textsc{XMark} achieves a BA comparable to MPAC.

\begin{table}
\centering
% \vspace{-0.8em}
\caption{Results on the machine translation task using WMT14 German-to-English dataset.}
\label{tab:translation}
\scalebox{0.80}{
\begin{tabular}{lcccccc}
\toprule
\textbf{Method} & \textbf{BA$\uparrow$} & \textbf{PPL$\downarrow$}  & \textbf{R.-1$\uparrow$} & \textbf{R.-Lsum$\uparrow$} & \textbf{BLEU$\uparrow$} \\
\midrule
RSBH & $58.62$ & $7.60$ & $0.61$ & $0.58$ & $38.96$ \\
MPAC & $60.88$ & $7.54$ & $0.65$ & $0.61$ & $43.21$ \\
\rowcolor[rgb]{ .9, .9, .9} \textsc{XMark}& $64.62$ & $7.26$ &  $0.64$ & $0.61$ & 44.52 \\
\bottomrule
\end{tabular}
}
\end{table}

\textbf{Results on Machine Translation Task.} We further evaluate \textsc{XMark} on a downstream machine translation task. Specifically, we use the WMT14 German-to-English dataset~\cite{wmt14}. In addition to BA and PPL, we also report BSc., R.-1, R.-Lsum, and BLEU~\cite{bleu}, where BLEU serves as a standard task-specific metric for translation quality. The results under message length $b=16$, bias $\delta=2$, and generation length $T=150$ are summarized in Table~\ref{tab:translation}. We observe that \textsc{XMark} achieves the highest BLEU score of $44.52$, outperforming the strongest baseline MPAC, which obtains $43.21$. This suggests that \textsc{XMark} imposes the smallest negative impact on the functional utility of the translation task. Meanwhile, \textsc{XMark} also achieves the highest BA of $64.62\%$, showing that it provides the best trade-off between decodability and task performance.

\begin{table}
\centering
% \vspace{-0.8em}
\caption{Performance comparison with long messages ($b=64$) and long generated texts ($T=1000$) on the text completion task.}
\label{tab:longermessage_longertext_moremetrics}
\scalebox{0.80}{
\begin{tabular}{lcccccc}
\toprule
\textbf{Method} & \textbf{BA$\uparrow$} & \textbf{PPL$\downarrow$}  & \textbf{R.-1$\uparrow$} & \textbf{R.-Lsum$\uparrow$} & \textbf{BSc.$\uparrow$} \\
\midrule
RSBH & $84.47$ & $4.46$  & $0.36$ & $0.33$ & $0.8266$\\
MPAC & $85.81$ & $4.71$ & $0.35$ & $0.32$  & $0.8254$\\
\rowcolor[rgb]{ .9, .9, .9} \textsc{XMark}& $93.00$ & $4.50$ & $0.38$ & $0.35$ & $0.8304$ \\
\bottomrule
\end{tabular}
}
\end{table}

\textbf{Longer Message, Longer Text, and More Semantic Metrics.} We further evaluate all methods under a larger-scale setting with a long message length ($b=64$) and a long generated text length ($T=1000$) on the text completion task. In addition to BA and PPL, we also report BSc. as well as R.-1 and ROUGE-Lsum (R.-Lsum)~\cite{lin2004rouge}. The results are summarized in Table~\ref{tab:longermessage_longertext_moremetrics}. We can see that \textsc{XMark} achieves the highest BA by a clear margin, improving from $85.81\%$ with MPAC to $93.00\%$. At the same time, it also provides the best overall text quality, achieving the highest BSc., R.-1, and R.-Lsum, while maintaining a low PPL comparable to the baselines. These results show that \textsc{XMark} remains highly effective for long messages, while better preserving both the semantic meaning and lexical content of watermarked texts.

\begin{table}
\centering
% \vspace{-0.8em}
\caption{Comparison of runtime (in seconds) and BA under the default setting.}
\label{tab:runtime_comparison}
\scalebox{0.80}{
\begin{tabular}{lccc}
\toprule
\textbf{Method} & \textbf{Encoding} & \textbf{Decoding} & \textbf{BA}$\uparrow$ \\
\midrule
DepthW & $11.51$ & $16.16$ & $79.50$ \\
MPAC & $10.93$ & $0.06$ & $94.00$ \\
RSBH & $11.80$ & $8.45$ & $92.75$ \\
StealthInk & $11.47$ & $0.07$ & $85.00$ \\
% \midrule
\rowcolor[rgb]{ .9, .9, .9} \textsc{XMark}& $11.54$ & $0.08$ & $98.75$ \\
\bottomrule
\end{tabular}
}
\end{table}
\textbf{Runtime Analysis.} We report the average encoding and decoding time under the default setting, with results shown in Table~\ref{tab:runtime_comparison}. The encoding time of all methods is around $11.5$ seconds, suggesting that the overall latency is mainly dominated by the LLM inference process itself. This also shows that \textsc{XMark} introduces negligible additional overhead during generation. For decoding, \textsc{XMark} remains highly efficient, requiring only $0.08$ seconds, which is comparable to the fastest baseline, MPAC ($0.06$ seconds). By contrast, DepthW and RSBH are much slower, taking $16.16$ and $8.45$ seconds, respectively, even for the short message length of $b=8$, mainly because they rely on candidate enumeration. These results show that \textsc{XMark} effectively overcomes the computational inefficiency of prior methods while still achieving substantially higher BA ($98.75\%$).

\begin{table}[t]
  \centering
  \caption{Ablation study of \textsc{XMark} with different numbers of hash keys ($k$) and different token–shard mapping matrices: cTMM in \textsc{XMark} vs. TMM in \textsc{XMark}$^-$.}
  \scalebox{0.69}{
    \begin{tabular}{lcccc|cc}
    \toprule
    \multirow{2}[2]{*}{\textbf{Method}} & \multicolumn{2}{c}{$T=150$} & \multicolumn{2}{c|}{$T=250$} & \multirow{1}[2]{*}{\textbf{\makecell*[c]{Avg. \\ BA$\uparrow$}}} & \multirow{1}[2]{*}{\textbf{\makecell*[c]{Avg. \\ PPL$\downarrow$}}} \\
\cmidrule(r){2-3}\cmidrule(r){4-5}       
         & \textbf{BA}$\uparrow$    & \textbf{PPL}$\downarrow$   & \textbf{BA}$\uparrow$    & \textbf{PPL}$\downarrow$  \\
\midrule
    \textsc{LoSo}   &   $74.06$   &  $4.45$   & $81.38$  &  $4.46$  & $77.72$ & $4.46$ \\
    \midrule \midrule
    \textsc{XMark} ($k{=}2$)  &  $82.62$    &  $4.84$   & $88.00$  &  $4.60$  & $85.31$ & $4.72$ \\
    \textsc{XMark} ($k{=}3$)   &   $82.94$   &   $5.01$  & $89.38$  &  $4.97$ & $86.16$ & $4.99$ \\
    \textsc{XMark} ($k{=}4$)   &  $84.25$    &   $5.07$  &  $92.06$ &  $5.14$  & $88.16$ & $5.11$ \\
    \midrule \midrule
    \textsc{XMark}$^-$ ($k{=}2$)  &  $81.31$    &  $-$   &  $88.06$ &  $-$  & $84.69$ & $-$ \\
    \textsc{XMark}$^-$ ($k{=}3$)   &   $80.69$   &   $-$  & $87.75$ &  $-$ & $84.22$ & $-$ \\
    \textsc{XMark}$^-$ ($k{=}4$)   &  $80.31$    &   $-$  &  $89.56$ &  $-$  & $84.94$ & $-$ \\
    \bottomrule
    \end{tabular}%
    }
  \label{tab: ablation}%
  % \vspace{-10pt}
\end{table}%

\textbf{Ablation Study.} We conduct an ablation study on \textsc{XMark} by varying the number of hash keys $k$ from $1$ to $4$. Note that when $k=1$, \textsc{XMark} reduces to \textsc{LoSo}. By default, \textsc{XMark} employs the cTMM in decoding. For comparison, we also evaluate \textsc{XMark} with the \textsc{TMM} in decoding, denoted as \textsc{XMark}$^-$. Experiments are conducted on the text completion task with $b=32$, with $T$ set to $150$ and $250$, and the results are summarized in Table~\ref{tab: ablation}. We ignore the PPL results for \textsc{XMark}$^-$ as they are the same as \textsc{XMark}'s PPL. The findings are clear: as $k$ increases, \textsc{XMark} achieves consistently higher BA, since additional keys allow more observations to accurately estimate the mapping between each generated token and its originating shard during decoding. However, this also slightly increases PPL due to the mildly reduced green list ratio. Compared with \textsc{XMark}, \textsc{XMark}$^-$ exhibits marginally lower BA due to the over-counting of the unboosted shard in TMM, especially when the number of generated tokens is small. Overall, the results indicate that cTMM is a more effective token-shard mapping construction method for \textsc{XMark}.

\textbf{Additional Results.} Results on more widely-used public LLMs and datasets are provided in Appendix~\ref{apdx: Moreresults}. We also discuss how \textsc{XMark} handles \textit{false positive cases} in Appendix~\ref{apdx: falsepositive}.

\section{Conclusion}
We propose a novel multi-bit watermarking method, \textsc{XMark}, for LLM-generated texts. During encoding, \textsc{XMark} leverages distinct hash keys to generate multiple permutations of the vocabulary, from which an evergreen list is constructed by intersecting the green lists across permutations. Tokens in the evergreen list are then boosted to increase their sampling probability. During decoding, each token can update multiple shard observations under the cTMM constraint, thereby providing a more accurate estimation of the mapping between a generated token and its originating shard. This enhanced counting mechanism ensures a more reliable decoding process, especially when the number of tokens for decoding is limited. Extensive experiments on diverse LLM downstream tasks and settings demonstrate that \textsc{XMark} substantially improves decoding accuracy compared with other methods, while preserving the quality of watermarked texts.

\clearpage

\section*{Limitations}

We now discuss the limitations and potential future directions of our work. 

First, during the decoding process of \textsc{XMark}, each row of the cTMM has a size of $1 \times 2^d$, resulting in a total of $r \times 2^d$ shards available for token–shard mappings. In our experiments, we fix $d=2$. Increasing $d$ substantially expands the number of shard candidates, which in turn reduces the number of observations/counts for the token–shard mappings per shard, given a fixed $k$ and $T,$ and decreases the accuracy of estimation, thereby degrading the decoding accuracy. Adapting \textsc{XMark} to larger $d$ values is an important future direction, as increasing $d$ can significantly enhance the quality of watermarked text, as discussed in Remark~\ref{remark: them}. 

Second, although \textsc{XMark} demonstrates robustness against text editing attacks in our evaluation, its design does not explicitly incorporate mechanisms to enhance robustness against attacks. Introducing such mechanisms represents another promising avenue for future research.

Third, as discussed in Remark~\ref{remark: them}, the hyperparameter $k$ controls the trade-off between text quality and decoding accuracy. Although we conducted an ablation study on $k$, in practice, the optimal value still requires manual tuning. Designing an adaptive strategy to automatically determine the optimal $k$ is another valuable direction for future exploration.

Finally, in our experiments, all methods are evaluated under a fixed $\delta$. In fact, $\delta$ also plays an important role in balancing text quality and decoding accuracy. Specifically, a larger $\delta$ yields more reliable decoding but may degrade the quality of the generated text. Jointly optimizing \textsc{XMark} with respect to $\delta$ constitutes an additional important direction for future work.

\section*{Acknowledgments}
We thank the anonymous reviewers and the area
chair for their constructive comments.
The work of Jiahao, Rui, and Zikai was supported in part by the National Science Foundation under Grant No. 2511989.
This material is based upon work co-supported by the U.S. Department of Energy, Office of Science, Office of Advanced Scientific Computing Research under Contract No. DE-AC05-00OR22725. This manuscript has been co-authored by UT-Battelle, LLC under Contract No. DE-AC05-00OR22725 with the U.S. Department of Energy. The United States Government retains and the publisher, by accepting the article for publication, acknowledges that the United States Government retains a non-exclusive, paid-up, irrevocable, world-wide license to publish or reproduce the published form of this manuscript, or allow others to do so, for United States Government purposes. The Department of Energy will provide public access to these results of federally sponsored research in accordance with the DOE Public Access Plan (http://energy.gov/downloads/doe-public-access-plan).

\bibliography{custom}

\clearpage

\appendix

\section{Appendix}
\subsection{Notation Table}

\begin{table}[t]
    \centering
    \caption{Notation table.}
    \label{tab: notation}
    \renewcommand{\arraystretch}{1.15}
    \scalebox{0.77}{
    \begin{tabular}{ll}
        \toprule
        \textbf{Symbol} & \textbf{Description} \\
        \midrule
        $\gamma$ & The green list ratio \\
        $\mathbf{m}$ & The binary multi-bit message \\
        $b$ & The length of message $\mathbf{m}$ \\
        $f$ & The large language model \\
        $\mathbf{x}_p$ & The input prompt \\
        $\mathbf{x}_s$ & The suspect text \\
        $x_t$ & The $t$-th token generated by $f$ \\
        $\mathcal{V}$ & The vocabulary of $f$ \\
        $\mathbf{x}_g$ & The generated text of $f$ without watermark \\
        $\mathbf{x}_g^\prime$ & The generated text of $f$ with watermark \\
        $T$ & The generation step of $f$ and length of $\mathbf{x}_g$ \\        
        $\mathcal{G}$ & The green list \\
        $\mathcal{S}$ & The token shard \\
        $\delta$ & The watermarking bias \\
        $\mathtt{Enc} (\cdot)$ & The encoding function \\
        $\mathtt{Dec} (\cdot)$ & The decoding function \\
        $\mathbf{m}^\prime$ & The decoded message via $\mathtt{Dec} (\cdot)$ \\
        $\mathtt{Quality}(\cdot)$ & The quality function \\
        $\mathtt{PPL}(\cdot)$ & The conditional perplexity \\
        $\tau$ & The tolerance margin on text quality\\
        $\mathbf{k}$ & The secret hash key \\
        $k$ & The number of hash keys \\
        $\mathtt{Hash}(\cdot)$ & The Hash function \\
        $\mathbf{s}$ & The pseudo-random seed generated by Hash \\
        $r$ & The number of blocks \\
        $d$ & The number of bits in a message block \\
        $\mathbf{A}$ & The token-shard mapping matrix \\
        $p$ & The decimal value of a message block \\
        $\mathcal{E}$ & The evergreen list \\
        \bottomrule
    \end{tabular}
    }
\end{table}
We present the detailed notation table in Table~\ref{tab: notation} for the reader's convenience.

\subsection{Hardware Settings.}
All experiments were carried out on a self-managed Linux-based computing cluster running Ubuntu 20.04.6 LTS. The cluster is equipped with eight NVIDIA RTX A6000 GPUs (each with 48 GB of memory) and AMD EPYC 7763 CPUs featuring 64 cores. Model inference leveraged GPU acceleration extensively. In total, the experiments accumulated roughly two weeks of GPU compute time.

\subsection{Related Work} \label{apdx: related_work}

\textbf{Zero-bit Watermarking for LLMs.} The first zero-bit watermarking approach, KGW, was proposed by~\citet{kgw}. At each generation step $t$, after the language model $f$ produces the logit vector for the next token, a pseudo-random seed is deterministically derived using a hash function that takes as input a hash key $\mathbf{k}$ and the previously generated token $x_{t-1}$. This seed governs a fixed permutation and subsequent partitioning of the vocabulary $\mathcal{V}$ into two disjoint subsets: the green list $\mathcal{G}$ and the red list $\mathcal{R}$. A positive bias $\delta$ is then added to the logits corresponding to tokens in $\mathcal{G}$, increasing their likelihood under the $\mathtt{softmax}(\cdot)$ distribution. While a larger $\delta$ strengthens the watermark signal, it also amplifies the distortion in the token sampling distribution, potentially degrading the fluency and naturalness of the generated text. Repeating this process at every generation step produces a watermarked sequence $\mathbf{x}_g^\prime$. For verification, given a suspect text $\mathbf{x}_s$, the decoder reconstructs the green list for each token position using the same key and hashing procedure, and then conducts a statistical $z$-test to evaluate whether the observed proportion of tokens from $\mathcal{G}$ significantly exceeds the expected baseline, thereby determining the presence of the watermark.
 %, to test if $\mathbf{x}_g'$ is watermarked. 
% test the hypothesis: \textit{`` this text was generated by an LLM''}. 
A statistically significant excess over the expected frequency indicates the presence of the watermark.
Several subsequent works have improved this zero-bit watermarking method to achieve higher decoding accuracy and better utility of the watermarked text~\cite{kirchenbauer2024on, kuditipudi2024robust, wang2025morphmark, postmark, giboulot2024watermax, liu2024a, piet2025markmywords, christ2024undetectable, munyer2024deeptextmark}. 

\textbf{Multi-bit Watermarking for LLMs.} While zero-bit watermarking enables text verification, it is insufficient for traceable watermarking, motivating the development of multi-bit watermarking methods. CycleShift~\cite{cycleshift} cyclically shifts vocabulary permutations according to the embedded message and biases the tokens within the resulting green list. However, overlapping shifts can introduce interference among message bits, reducing distinctiveness and weakening statistical separability~\cite{stealthink}. DepthW~\cite{depthw} directly encodes the message as input to the hash function and further sets a dark green list inside of the green list, to which stronger perturbations are applied. Despite their effectiveness, both CycleShift and DepthW rely on brute-force search over all possible message candidates, rendering them impractical for long messages.

To improve decoding efficiency, MPAC~\cite{mpac} partitions the message into multiple blocks and uses each block's value to guide green list construction. However, as discussed in this paper, MPAC substantially degrades text quality. RSBH~\cite{rsbh} enhances MPAC by incorporating the message block value into the hash seed computation and adopting a larger green list ratio $\gamma = 0.5$. This design improves text utility but increases decoding complexity exponentially with $2^d$. More recently, StealthInk~\cite{stealthink} improves text quality by directly amplifying the sampling probabilities of high-logit tokens while suppressing low-logit ones, thereby preserving fluency. Nevertheless, this approach weakens the watermark signal and consequently reduces decoding accuracy.

\begin{algorithm}[t]
\normalsize
\caption{The Decoder of \textsc{XMark}}
\label{alg: XMark_decoder}
\SetKwInOut{Input}{Input}
\SetKwInOut{Output}{Output}
\Input{Vocabulary $\mathcal{V}$, hash keys $\{ \mathbf{k}_0, \mathbf{k}_1, \dots, \mathbf{k}_{k-1} \}$, number of blocks $r$, length of message block $d$, suspect text $\mathbf{x}^\prime$ of length $T$}
\Output{Decoded message $\mathbf{m}^\prime$}
Initialize $\mathbf{A} \leftarrow \mathbf{0}^{r \times 2^d}$ 

    \For{$t=3$ \KwTo $T$}{
        % $x_{t-1}^\prime, x_{t-2}^\prime \leftarrow \mathbf{x}_{s,t{-}1}, \mathbf{x}_{s,t{-}2}$ \\
        
        $i \leftarrow (x_{t-2}^\prime + x_{t-1}^\prime) \mod r$ \\
 
        $\mathbf{cc} \leftarrow \mathbf{0}^{1 \times 2^d}$
        
        \For{$j\leftarrow0$ \KwTo $k-1$}{
            $\mathbf{s}_j \leftarrow \mathtt{Hash}( x_{t-2}^\prime, x_{t-1}^\prime, \mathbf{k}_j)$ \\
            $\mathcal{V}^\prime \leftarrow \mathtt{permute}(\mathcal{V}, \mathbf{s}_j)$ \\
            $\mathcal{S}_{j,0}, \dots, \mathcal{S}_{j,2^d-1} \leftarrow \mathtt{part}(\mathcal{V}^\prime, 2^d)$ \\
            % \For{$i = 1$ \KwTo $b$}{
                \If{$x_t^\prime \in \mathcal{S}_{j, u}$}{
                    $\mathbf{cc}[u] \leftarrow \mathbf{cc}[u]+ 1$ \\
                    % \textbf{break}
                % }
            }
        }

        $\mathbf{update}\leftarrow \mathtt{sign}(\mathbf{cc})$ \label{algline: sign}
        
        $\mathbf{A}[i,:] \leftarrow \mathbf{A}[i,:] + \mathbf{update}$
        }

\For{$i\leftarrow0$ \KwTo $r-1$}{
        $p^\prime_i \leftarrow \arg \min \mathbf{A}[i:]$ \\

        $\mathbf{m}_i^\prime \leftarrow \mathtt{binary} (p^\prime_i)$
        }

\textbf{Return} $\mathbf{m}^\prime = \mathtt{concat}(\mathbf{m_0^\prime}, \dots, \mathbf{m}_{r-1}^\prime)$
\end{algorithm}

\subsection{Decoder Algorithm} \label{apdx: decoder_alg}

We present the detailed decoding procedure of \textsc{XMark} in Algorithm~\ref{alg: XMark_decoder}. Note that we start from the third token during decoding because computing the block index and hash seed requires $x_{t-2}^\prime$ and $x_{t-1}^\prime$. 
% which are the last two tokens of the user prompt (unavailable in practice) during encoding. 
In Line~\ref{algline: sign}, the function $\mathtt{sign}(\cdot)$ operates element-wise, setting any value greater than $0$ to $1$ while keeping $0$ unchanged, thereby producing the constrained token-shard mapping matrix (cTMM).

\subsection{System Prompt Settings} \label{apdx: systemprompt}

Recall that in our work, we use the LLaMA-$2$-$7$B-chat model~\cite{llama2} for story generation and text summarization tasks. The chat prompt for story generation is designed to encourage coherent and imaginative story generation and is defined as follows:
\begin{quote}
\texttt{[System]} \textit{You are a helpful assistant that writes engaging and coherent stories.}

\texttt{[User]} \textit{Please write a detailed and imaginative short story based on the following prompt: [prompt].}
\end{quote}

For the text summarization task, the chat prompt is designed to instruct the model to generate concise and faithful summaries given an input article, using the following format:
\begin{quote}
\texttt{[System]} \textit{You are a helpful assistant specialized in summarization. You take a document and write a concise, faithful summary.}

\texttt{[User]} \textit{Please summarize the following article in a few sentences: [article].}
\end{quote}

\begin{table*}[ht]
  \centering
  \caption{
Performance comparison on text completion tasks with message length $b=16$ under different numbers of available tokens $T \in \{150, 200, 250, 300\}$ on the Essays and OpenGen datasets.
}
  \scalebox{0.75}{
    \begin{tabular}{c|l|cc|cc|cc|cc|cc}
    \toprule
    % \footnotesize
    \multirow{2}[2]{*}{\textbf{\makecell*[c]{Task \\ (Dataset)}}} 
    & \multirow{2}[2]{*}{\textbf{Method}} 
    & \multicolumn{2}{c|}{$T=150$}   
    & \multicolumn{2}{c|}{$T=200$}   
    & \multicolumn{2}{c|}{$T=250$} 
    & \multicolumn{2}{c|}{$T=300$} 
    & \multirow{2}[2]{*}{\textbf{\makecell*[c]{Avg. \\ BA$\uparrow$}}} 
    & \multirow{2}[2]{*}{\textbf{\makecell*[c]{Avg. \\ PPL$\downarrow$}}} \\
    \cmidrule(r){3-4}\cmidrule(r){5-6}\cmidrule(r){7-8}\cmidrule(r){9-10}
          &       & \textbf{BA}$\uparrow$    & \textbf{PPL}$\downarrow$   
          & \textbf{BA}$\uparrow$    & \textbf{PPL}$\downarrow$   
          & \textbf{BA}$\uparrow$    & \textbf{PPL}$\downarrow$
          & \textbf{BA}$\uparrow$    & \textbf{PPL}$\downarrow$ \\
    \midrule
    \multirow{4}[2]{*}{\makecell*[c]{Text Completion \\ (Essays)}} 
    & StealthInk       & $71.62$ & $4.12$ & $75.00$ & $3.90$ & $78.88$ & $3.75$ & $79.25$ & $3.66$ & $76.19$ & $3.86$ \\
    & MPAC             & $85.88$ & $5.37$ & $89.12$ & $5.09$ & $91.75$ & $4.93$ & $92.88$ & $4.78$ & $89.91$ & $5.04$ \\
    & RSBH             & $80.12$ & $4.97$ & $85.00$ & $4.80$ & $89.75$ & $4.66$ & $91.00$ & $4.53$ & $86.47$ & $4.74$ \\
    & \cellcolor[rgb]{ .9, .9, .9}\textsc{XMark} & \cellcolor[rgb]{ .9, .9, .9}$92.62$ & \cellcolor[rgb]{ .9, .9, .9}$4.88$ & \cellcolor[rgb]{ .9, .9, .9}$95.62$ & \cellcolor[rgb]{ .9, .9, .9}$4.68$ & \cellcolor[rgb]{ .9, .9, .9}$97.12$ & \cellcolor[rgb]{ .9, .9, .9}$4.50$ & \cellcolor[rgb]{ .9, .9, .9}$97.75$ & \cellcolor[rgb]{ .9, .9, .9}$4.40$ & \cellcolor[rgb]{ .9, .9, .9}$95.78$ & \cellcolor[rgb]{ .9, .9, .9}$4.62$ \\
    \midrule \midrule
    \multirow{4}[2]{*}{\makecell*[c]{Text Completion \\ (OpenGen)}} 
    & StealthInk       & $72.75$ & $4.11$ & $75.75$ & $3.98$ & $77.88$ & $4.04$ & $80.12$ & $3.85$ & $76.63$ & $3.99$ \\
    & MPAC             & $84.88$ & $5.01$ & $89.25$ & $4.98$ & $90.38$ & $4.97$ & $92.62$ & $4.84$ & $89.28$ & $4.95$ \\
    & RSBH             & $77.00$ & $4.73$ & $81.75$ & $4.75$ & $88.25$ & $4.72$ & $87.12$ & $4.56$ & $83.53$ & $4.69$ \\
    & \cellcolor[rgb]{ .9, .9, .9}\textsc{XMark} & \cellcolor[rgb]{ .9, .9, .9}$90.62$ & \cellcolor[rgb]{ .9, .9, .9}$4.70$ & \cellcolor[rgb]{ .9, .9, .9}$91.12$ & \cellcolor[rgb]{ .9, .9, .9}$4.68$ & \cellcolor[rgb]{ .9, .9, .9}$95.25$ & \cellcolor[rgb]{ .9, .9, .9}$4.66$ & \cellcolor[rgb]{ .9, .9, .9}$95.88$ & \cellcolor[rgb]{ .9, .9, .9}$4.67$ & \cellcolor[rgb]{ .9, .9, .9}$93.22$ & \cellcolor[rgb]{ .9, .9, .9}$4.68$ \\
    \bottomrule
    \end{tabular}%
    }
  \label{tab: essays_opengen_b16}%
\end{table*}

\begin{table*}[ht]
  \centering
  \caption{
Performance comparison on text completion tasks with message length $b=32$ under different numbers of available tokens $T \in \{150, 200, 250, 300\}$ on the Essays and OpenGen datasets.
}
  \scalebox{0.75}{
    \begin{tabular}{c|l|cc|cc|cc|cc|cc}
    \toprule
    % \footnotesize
    \multirow{2}[2]{*}{\textbf{\makecell*[c]{Task \\ (Dataset)}}} 
    & \multirow{2}[2]{*}{\textbf{Method}} 
    & \multicolumn{2}{c|}{$T=150$}   
    & \multicolumn{2}{c|}{$T=200$}   
    & \multicolumn{2}{c|}{$T=250$} 
    & \multicolumn{2}{c|}{$T=300$} 
    & \multirow{2}[2]{*}{\textbf{\makecell*[c]{Avg. \\ BA$\uparrow$}}} 
    & \multirow{2}[2]{*}{\textbf{\makecell*[c]{Avg. \\ PPL$\downarrow$}}} \\
    \cmidrule(r){3-4}\cmidrule(r){5-6}\cmidrule(r){7-8}\cmidrule(r){9-10}
          &       & \textbf{BA}$\uparrow$    & \textbf{PPL}$\downarrow$   
          & \textbf{BA}$\uparrow$    & \textbf{PPL}$\downarrow$   
          & \textbf{BA}$\uparrow$    & \textbf{PPL}$\downarrow$
          & \textbf{BA}$\uparrow$    & \textbf{PPL}$\downarrow$ \\
    \midrule
    \multirow{4}[2]{*}{\makecell*[c]{Text Completion \\ (Essays)}} 
    & StealthInk        & $65.94$ & $4.17$ & $67.76$ & $3.96$ & $71.12$ & $3.84$ & $72.12$ & $3.74$ & $69.24$ & $3.93$ \\
    & MPAC              & $77.81$ & $5.03$ & $80.38$ & $4.89$ & $80.94$ & $4.76$ & $83.19$ & $4.65$ & $80.58$ & $4.83$ \\
    & RSBH              & $75.06$ & $5.07$ & $80.38$ & $4.77$ & $82.00$ & $4.67$ & $84.06$ & $4.54$ & $80.38$ & $4.76$ \\
    & \cellcolor[rgb]{ .9, .9, .9}\textsc{XMark}  & \cellcolor[rgb]{ .9, .9, .9}$82.38$ & \cellcolor[rgb]{ .9, .9, .9}$4.89$ & \cellcolor[rgb]{ .9, .9, .9}$85.38$ & \cellcolor[rgb]{ .9, .9, .9}$4.65$ & \cellcolor[rgb]{ .9, .9, .9}$89.31$ & \cellcolor[rgb]{ .9, .9, .9}$4.44$ & \cellcolor[rgb]{ .9, .9, .9}$91.25$ & \cellcolor[rgb]{ .9, .9, .9}$4.32$ & \cellcolor[rgb]{ .9, .9, .9}$87.08$ & \cellcolor[rgb]{ .9, .9, .9}$4.58$ \\
    \midrule \midrule
    \multirow{4}[2]{*}{\makecell*[c]{Text Completion \\ (OpenGen)}} 
    & StealthInk        & $67.12$ & $4.22$ & $67.88$ & $4.03$ & $69.94$ & $4.07$ & $71.69$ & $4.01$ & $69.16$ & $4.08$ \\
    & MPAC              & $76.06$ & $4.97$ & $79.31$ & $4.91$ & $82.69$ & $4.97$ & $84.94$ & $4.99$ & $80.75$ & $4.96$ \\
    & RSBH              & $70.88$ & $4.67$ & $76.06$ & $4.82$ & $78.62$ & $4.62$ & $77.56$ & $4.71$ & $75.78$ & $4.71$ \\
    & \cellcolor[rgb]{ .9, .9, .9}\textsc{XMark} & \cellcolor[rgb]{ .9, .9, .9}$79.50$ & \cellcolor[rgb]{ .9, .9, .9}$4.81$ & \cellcolor[rgb]{ .9, .9, .9}$83.88$ & \cellcolor[rgb]{ .9, .9, .9}$4.52$ & \cellcolor[rgb]{ .9, .9, .9}$85.81$ & \cellcolor[rgb]{ .9, .9, .9}$4.75$ & \cellcolor[rgb]{ .9, .9, .9}$90.62$ & \cellcolor[rgb]{ .9, .9, .9}$4.68$ & \cellcolor[rgb]{ .9, .9, .9}$84.95$ & \cellcolor[rgb]{ .9, .9, .9}$4.69$ \\
    \bottomrule
    \end{tabular}%
    }
  \label{tab: essays_opengen_b32}%
\end{table*}

\subsection{Discussion on False Positive Cases} \label{apdx: falsepositive}

False positives are a common issue across all multi-bit watermarking methods. 
Specifically, even when applied to unwatermarked text, a decoder may still output a message that could be mistakenly interpreted as a valid watermark. 
However, this problem has received little attention in the existing literature. 
We observe that our method, \textsc{XMark}, is inherently robust to false positives. 
Recall that during decoding, \textsc{XMark} leverages the cTMM to recover the embedded message. 
Our recovery strategy identifies the shard with the fewest token--shard mapping counts, and this statistical property naturally serves as an indicator of false positives: if no shard exhibits a noticeably small count, the text is likely unwatermarked. 
In practice, one can set a threshold based on the entropy of the token--shard count distribution for each message block. 
If the entropy exceeds this threshold, the decoder concludes that the text is unwatermarked; otherwise, it proceeds with message recovery as usual.

We further conduct empirical experiments to evaluate \textsc{XMark}'s ability to detect false positives, and compare it with CycleShift, MPAC, and StealthInk under the setting of $b=8$, $T=150$, and $\delta=2$. For each baseline, we follow the detection metric used in its original paper, such as the $p$-value for CycleShift and the $z$-score for MPAC, and vary the corresponding decision threshold to measure detection performance. For \textsc{XMark}, given the well-updated cTMM $\mathbf{A}^T$, we compute the standard deviation of each row (corresponding to each message block) and then take the average. We report the TPR at a fixed $10\%$ FPR, together with the F$1$ score, in Table~\ref{tab:fpr_b}. Note that RSBH and DepthW are not included in this comparison, as their original papers do not provide explicit false-positive detection modules or statistical thresholds for unwatermarked text detection. From the results, \textsc{XMark} achieves strong detection performance, with a TPR of $94\%$ at $10\%$ FPR and an F$1$ score of $0.92$. It substantially outperforms CycleShift and StealthInk, and remains competitive with MPAC, which achieves slightly higher detection metrics. We note that MPAC benefits from a stricter green list ratio ($\gamma=0.25$), which produces a stronger detection signal but typically comes at the cost of text quality. Overall, these results show that \textsc{XMark} provides a favorable balance between detection reliability and generation quality.
\begin{table*}[t]
\centering
\caption{False positive detection performance measured by TPR at $10\%$ FPR (TPR@$10\%$FPR) and F$1$ score.}
\label{tab:fpr_b}
\scalebox{0.8}{
\begin{tabular}{c|c|c|c}
\toprule
\textbf{Method} & \textbf{Detection Mechanism and Threshold} & \textbf{TPR@$10\%$FPR $\uparrow$} & \textbf{F1 Score $\uparrow$}\\
\midrule
CycleShift & $p$-value ($p<2\mathrm{e}^{-2}$) & $80.00$ & $0.85$ \\
MPAC & $z$-score ($z > 4.10$) & $98.00$ & $0.94$ \\
StealthInk & $p$-value ($p<8\mathrm{e}^{-4}$) & $60.00$ & $0.70$ \\
\rowcolor[rgb]{.9, .9, .9}\textsc{XMark} & Standard Deviation ($\sigma > 3.30$) & $94.00$ & $0.92$ \\
\bottomrule
\end{tabular}
}
\end{table*}

\subsection{More Results} \label{apdx: Moreresults}

\begin{table*}[t]
  \centering
  \caption{Performance comparison of MPAC and \textsc{XMark} on Qwen$2.5$-$7$B and LLaMA-$3.1$-$8$B models with message lengths $b \in \{8, 16\}$ and number of available tokens $T \in \{150, 200, 250, 300\}$.}
  \scalebox{0.73}{
    \begin{tabular}{c|l|l|cc|cc|cc|cc|cc}
    \toprule
    \footnotesize
    \multirow{2}[2]{*}{\textbf{\makecell*[c]{Message \\ Length}}} &
    \multirow{2}[2]{*}{\textbf{Model}} &
    \multirow{2}[2]{*}{\textbf{Method}} &
    \multicolumn{2}{c|}{$T=150$} &
    \multicolumn{2}{c|}{$T=200$} &
    \multicolumn{2}{c|}{$T=250$} &
    \multicolumn{2}{c|}{$T=300$} &
    \multirow{2}[2]{*}{\textbf{\makecell*[c]{Avg. \\ BA$\uparrow$}}} &
    \multirow{2}[2]{*}{\textbf{\makecell*[c]{Avg. \\ PPL$\downarrow$}}} \\
    \cmidrule(lr){4-5}\cmidrule(lr){6-7}\cmidrule(lr){8-9}\cmidrule(lr){10-11}
          &   &    &
          \textbf{BA}$\uparrow$ & \textbf{PPL}$\downarrow$ &
          \textbf{BA}$\uparrow$ & \textbf{PPL}$\downarrow$ &
          \textbf{BA}$\uparrow$ & \textbf{PPL}$\downarrow$ &
          \textbf{BA}$\uparrow$ & \textbf{PPL}$\downarrow$ \\
    \midrule
    \multirow{4}[4]{*}{$b=8$} &
    \multirow{2}[2]{*}{Qwen$2.5$-$7$B} &
    MPAC            & $96.50$ & $8.25$ & $98.00$ & $7.69$ & $98.50$ & $7.57$ & $98.50$ & $7.66$ & $97.88$ & $7.79$ \\
    & & \cellcolor[rgb]{ .9, .9, .9}\textsc{XMark} & \cellcolor[rgb]{ .9, .9, .9}$97.75$ & \cellcolor[rgb]{ .9, .9, .9}$7.66$ & \cellcolor[rgb]{ .9, .9, .9}$100.00$ & \cellcolor[rgb]{ .9, .9, .9}$7.19$ & \cellcolor[rgb]{ .9, .9, .9}$100.00$ & \cellcolor[rgb]{ .9, .9, .9}$7.07$ & \cellcolor[rgb]{ .9, .9, .9}$100.00$ & \cellcolor[rgb]{ .9, .9, .9}$7.00$ & \cellcolor[rgb]{ .9, .9, .9}$99.44$ & \cellcolor[rgb]{ .9, .9, .9}$7.23$ \\
    \cmidrule(lr){2-13}
    & \multirow{2}[2]{*}{LLaMA-$3.1$-$8$B} &
    MPAC            & $97.25$ & $6.55$ & $98.25$ & $6.44$ & $98.75$ & $6.38$ & $99.50$ & $6.33$ & $98.44$ & $6.43$ \\
    & & \cellcolor[rgb]{ .9, .9, .9}\textsc{XMark} & \cellcolor[rgb]{ .9, .9, .9}$100.00$ & \cellcolor[rgb]{ .9, .9, .9}$6.37$ & \cellcolor[rgb]{ .9, .9, .9}$99.75$ & \cellcolor[rgb]{ .9, .9, .9}$6.05$ & \cellcolor[rgb]{ .9, .9, .9}$100.00$ & \cellcolor[rgb]{ .9, .9, .9}$5.98$ & \cellcolor[rgb]{ .9, .9, .9}$100.00$ & \cellcolor[rgb]{ .9, .9, .9}$5.94$ & \cellcolor[rgb]{ .9, .9, .9}$99.94$ & \cellcolor[rgb]{ .9, .9, .9}$6.09$ \\
    \midrule \midrule
    \multirow{4}[4]{*}{$b=16$} &
    \multirow{2}[2]{*}{Qwen$2.5$-$7$B} &
    MPAC           & $91.12$ & $7.97$ & $91.00$ & $7.90$ & $94.75$ & $7.54$ & $95.38$ & $7.55$ & $93.06$ & $7.74$ \\
    & & \cellcolor[rgb]{ .9, .9, .9}\textsc{XMark} & \cellcolor[rgb]{ .9, .9, .9}$96.38$ & \cellcolor[rgb]{ .9, .9, .9}$7.62$ & \cellcolor[rgb]{ .9, .9, .9}$97.12$ & \cellcolor[rgb]{ .9, .9, .9}$7.07$ & \cellcolor[rgb]{ .9, .9, .9}$97.75$ & \cellcolor[rgb]{ .9, .9, .9}$6.92$ & \cellcolor[rgb]{ .9, .9, .9}$99.25$ & \cellcolor[rgb]{ .9, .9, .9}$6.76$ & \cellcolor[rgb]{ .9, .9, .9}$97.63$ & \cellcolor[rgb]{ .9, .9, .9}$7.09$ \\
    \cmidrule(lr){2-13}
    & \multirow{2}[2]{*}{LLaMA-$3.1$-$8$B} &
    MPAC           & $91.00$ & $7.03$ & $92.25$ & $6.53$ & $95.00$ & $6.69$ & $96.62$ & $6.39$ & $93.72$ & $6.66$ \\
    & & \cellcolor[rgb]{ .9, .9, .9}\textsc{XMark} & \cellcolor[rgb]{ .9, .9, .9}$95.50$ & \cellcolor[rgb]{ .9, .9, .9}$6.26$ & \cellcolor[rgb]{ .9, .9, .9}$98.00$ & \cellcolor[rgb]{ .9, .9, .9}$6.07$ & \cellcolor[rgb]{ .9, .9, .9}$97.88$ & \cellcolor[rgb]{ .9, .9, .9}$5.93$ & \cellcolor[rgb]{ .9, .9, .9}$98.62$ & \cellcolor[rgb]{ .9, .9, .9}$6.01$ & \cellcolor[rgb]{ .9, .9, .9}$97.50$ & \cellcolor[rgb]{ .9, .9, .9}$6.07$ \\
    \bottomrule
    \end{tabular}%
  }
  \label{tab:qwen}
\end{table*}

\begin{table*}[ht]
    \centering
    \caption{Example of generated texts of \textsc{XMark}.}
    \label{tab:generated_text}
    \textbf{Prompt}: \textit{MIDDLETOWN, NJ - The Middletown Township Public Library and the Township of Middletown are pleased to announce the appointment of Kim Rinaldi as Provisional Library Director, beginning on February 24, 2014. [...] choice for the residents of the Township.} \\
    \renewcommand{\arraystretch}{1.2}
    \scalebox{0.7}{
    \begin{tabular}{l p{15cm} c}  % 控制Prompt和Text的宽度
        \toprule
        \textbf{Method} & \textbf{Generated Text} & \textbf{PPL}$\downarrow$  \\
        \midrule
        Nonwatermarked & The Middletown Board of Trustees interviewed three highly qualified candidates for the Library Director position. The Board of Trustees voted to appoint Ms. Rialdi to the position of Provisional Library director at a meeting held on February 10, 2004.
The MTPL is a member of the Monmouth County Library & $3.17$  \\ \\
        
        \cellcolor[rgb]{ .9, .9, .9}\textsc{XMark} & \cellcolor[rgb]{ .9, .9, .9}The provisional library director contract will be ratified by the Middletown Township Public Library Board of Trustees on Tuesday, February 25, 2024. Ms Rinaldi will begin on Monday, February 24th. Previous MTPL Receives 2013 & \cellcolor[rgb]{ .9, .9, .9}$3.92$ \\
        \bottomrule
    \end{tabular}
    } \\ \vspace{10pt}
    \textbf{Prompt}: \textit{Belying expectations, Prasar Bharti has earned only Rs 58.19 crore (Rs 581.9 million) as revenue during the Commonwealth Games [...] Rs 90,000 for the same in the opening of the CWG.} \\
    \scalebox{0.7}{
    \begin{tabular}{l p{15cm} c}  % 控制Prompt和Text的宽度
        \toprule
        \textbf{Method} & \textbf{Generated Text} & \textbf{PPL}$\downarrow$  \\
        \midrule
        Nonwatermarked & The advertisements were supposed to be telecasted live and the revenue was supposed to be earned in real time. But after the complaints, the channel was not allowed to telecast advertisements during closing ceremony.
The opening ceremony was telecasted live on DD and was deferred live for the evening slot to accommodate advertising. & $4.61$  \\ \\
        
        \cellcolor[rgb]{ .9, .9, .9}\textsc{XMark} & \cellcolor[rgb]{ .9, .9, .9}But the channel had to forego the rate hike as advertisers did not want to be associated with a sports event, sources said.
Prasar Bharati's revenue from the last CWG Games was Rs 59.19 cr (Rs 403.8 million), a sharp drop from Rs & \cellcolor[rgb]{ .9, .9, .9}$5.49$ \\
        \bottomrule
    \end{tabular}
    }
\end{table*}

\textbf{More Results on Additional Text Completion Tasks.}
We further evaluate MPAC, RSBH, StealthInk, and our proposed \textsc{XMark} on the text completion task using the OpenGen~\cite{opengen} and Essays~\cite{essays} datasets. Results for message lengths $b=16$ and $b=32$ with $T \in \{150, 200, 250, 300\}$ are summarized in Table~\ref{tab: essays_opengen_b16} and Table~\ref{tab: essays_opengen_b32}, respectively.
Across all $T$ settings, \textsc{XMark} consistently achieves higher BA than the compared methods. When $b=16$, on Essays and OpenGen, \textsc{XMark} attains the highest average BA of $95.78\%$ and $93.22\%$, yielding gains of $+5.87\%$ and $+3.94\%$ over the second-best method (MPAC). In addition to BA improvements, \textsc{XMark} also demonstrates clear PPL reductions, achieving PPL values of $4.62$ and $4.68$, respectively. When $b=32$, the advantage of \textsc{XMark} becomes even more pronounced, as the BA gaps over MPAC further widen to $+6.50\%$ on Essays and $+4.20\%$ on OpenGen.

\textbf{More Results on Additional LLMs.}
We further evaluate our method on two widely used language models: Qwen$2.5$-$7$B~\cite{qwen2} and LLaMA-$3.1$-$8$B~\cite{grattafiori2024llama}. The BA and PPL results with message lengths $b \in \{8, 16\}$ and $T \in \{150, 200, 250, 300\}$ are reported in Table~\ref{tab:qwen}. For PPL evaluation, we employ the larger models Qwen$2.5$-$32$B and LLaMA-$3.1$-$70$B to obtain more accurate text quality measurements.
As shown, \textsc{XMark} consistently achieves higher BA and lower PPL than MPAC across all configurations, confirming its strong generalization ability across different LLM architectures. Specifically, when $b=8$, \textsc{XMark} attains nearly perfect BA while maintaining average PPLs of $7.23$ and $6.09$ on Qwen$2.5$-$7$B and LLaMA-$3.1$-$8$B, respectively. When $b$ increases to $16$, \textsc{XMark} still delivers very high BA ($97.63\%$ and $97.50\%$), outperforming MPAC by $+4.57\%$ and $+3.78\%$, respectively. Moreover, \textsc{XMark} achieves lower PPLs of $7.09$ and $6.07$, corresponding to improvements of $+0.65$ and $+0.59$ compared with MPAC.

\subsection{Generated Text Example}
Table~\ref{tab:generated_text} presents examples of generated texts produced by \textsc{XMark} for a given prompt, with watermarking bias $\delta=2$ and message length $b=16$.

% \clearpage
\subsection{Proof of the Expected Green List Size of \textsc{XMark}} \label{apdx: proof}

\paragraph{Setup and notation.}
Let $\mathcal V$ denote the vocabulary, and fix integers $d \ge 1$ and $k \ge 1$, representing the length of the message block and the number of hash keys, respectively. 
For each hash key, we independently (i) generate a uniformly random permutation of $\mathcal V$, and (ii) partition the permuted vocabulary into $2^d$ disjoint shards of (nearly) equal size\footnote{We assume that $|\mathcal V|$ is divisible by $2^d$; otherwise, there is an $O(1/|\mathcal V|)$ discrepancy due to the near-equal partition.}. For a given key $j \in \{0, \dots, k-1\}$, we define the \emph{green list} $\mathcal G_j$ as the complement of one designated shard. Consequently, a uniformly chosen token belongs to $\mathcal G_j$ with probability $1 - 2^{-d}$. 
The \emph{evergreen list} is then defined as the intersection 
\[
\mathcal E \;=\; \bigcap_{j=0}^{k-1} \mathcal G_j.
\]
In the following, we prove that the expected size of $\mathcal E$ is $(1 - 2^{-d})^{k} |\mathcal V|.
$

\begin{proof}
For each token $v \in \mathcal V$, define the indicator variable
\[
X_v = \mathbf 1\!\{v \in \mathcal E\} = \prod_{j=0}^{k-1} \mathbf 1\!\{v \in \mathcal G_j\}.
\]
By construction and independence across keys, we have 
$\Pr[v \in \mathcal G_j] = 1 - 2^{-d},
$
and 
$\{\mathbf 1\{v \in \mathcal G_j\}\}_{j=0}^{k-1} \text{ are independent.}$
Therefore, we have
\begin{align*}
\mathbb E[X_v]
&= \Pr[v \in \bigcap_{j=0}^{k-1} \mathcal G_j] \\
&= \prod_{j=0}^{k-1} \Pr[v \in \mathcal G_j]
= (1 - 2^{-d})^{k}.
\end{align*}

Since $|\mathcal E| = \sum_{v \in \mathcal V} X_v$, by linearity of expectation, we have
\[
\mathbb E[|\mathcal E|]
= \sum_{v \in \mathcal V} \mathbb E[X_v]
= (1 - 2^{-d})^{k} |\mathcal V|.
\]
Dividing both sides by $|\mathcal V|$ gives
\[
\mathbb E[\gamma]
= \mathbb E\!\left[\frac{|\mathcal E|}{|\mathcal V|}\right]
= (1 - 2^{-d})^{k},
\]
which concludes the proof.
\end{proof}

\end{document}